%% file: main.tex
\documentclass{article} 

\usepackage{main,times}

\input{math_commands.tex}

\def\eqref#1{Eq.~(\ref{#1})}

\newcommand{\vnu}{\bm{\nu}}

\usepackage{hyperref}
\usepackage{url}

\usepackage{changepage}

\usepackage{booktabs}       
\usepackage{amsfonts}      
\usepackage{nicefrac}      
\usepackage{microtype}     
\usepackage{amssymb}
\usepackage{algorithm}
\usepackage{algpseudocode}
\usepackage{adjustbox}
\usepackage{lipsum}
\usepackage{array}
\usepackage{pgf-pie}
\usepackage{graphicx}
\usepackage{wrapfig}
\usepackage{amsthm}
\usepackage{thmtools, thm-restate}
\usepackage{mathtools}
\usepackage{subcaption}
\usepackage{multirow}
\usepackage{caption}
\usepackage{adjustbox} 
\usepackage{algpseudocode}
\usepackage{tabularx}
\newcolumntype{Y}{>{\raggedright\arraybackslash}X}

\algdef{SE}[WITH]{With}{EndWith}[1]{\textbf{with} #1:}{\textbf{end with}}

\usepackage{makecell}
\usepackage{multirow}        
\usepackage{siunitx}         
\usepackage{threeparttable} 
\usepackage{caption}
\usepackage{dsfont}

\newcommand{\ind}{\mathds{1}}
\newcommand{\real}{\hat p_N}

\definecolor{DeepGreen}{RGB}{0,83,0}
\definecolor{DeepBlue}{RGB}{0,50,120}

\definecolor{outerblue}{RGB}{30,90,255}
\definecolor{midorange}{RGB}{255,140,0}
\definecolor{innergreen}{RGB}{0,160,0}
\definecolor{deepviolet}{RGB}{150,0,180}

\newcommand{\appropto}{\mathrel{\vcenter{\offinterlineskip\halign{\hfil$##$\cr\propto\cr\noalign{\kern2pt}\sim\cr\noalign{\kern-2pt}}}}}

\usepackage{listings}
\lstdefinelanguage{prompt}{
  morestring=[b]",
  morecomment=[l]{\#},
  sensitive=true
}
\newcolumntype{C}[1]{>{\centering\arraybackslash}m{#1}}

\usepackage[table,dvipsnames]{xcolor}
\usepackage[most]{tcolorbox}
\newtcolorbox{takeawaybox}{
    enhanced,
    breakable,
    colback=midorange!5,
    colframe=midorange!55!black,
    boxrule=0.6pt,
    arc=1.5mm,
    borderline west={2.2pt}{0pt}{midorange!85!black},
    left=1.5mm,
    right=1.5mm,
    top=1.0mm,
    bottom=1.0mm,
    before skip=6pt,
    after skip=6pt,
}

\title{Does Uniform Discrete Diffusion Need Time?}

\iclrfinalcopy 

\author{
\makebox[0.30\textwidth][c]{\textbf{Chunsan Hong}\textsuperscript{1,2}}
\makebox[0.30\textwidth][c]{\textbf{Chieh-Hsin Lai}\textsuperscript{2}}
\makebox[0.30\textwidth][c]{\textbf{Satoshi Hayakawa}\textsuperscript{3}}
\\
\makebox[0.30\textwidth][c]{\textbf{Yuhta Takida\textsuperscript{2}}}
\makebox[0.30\textwidth][c]{\textbf{Jong Chul Ye\textsuperscript{1,$\dagger$}}}
\makebox[0.30\textwidth][c]{\textbf{Yuki Mitsufuji\textsuperscript{2,$\dagger$}}}
\\[1mm]
\makebox[0.30\textwidth][c]{\textsuperscript{1}KAIST}
\makebox[0.30\textwidth][c]{\textsuperscript{2}Sony Group Corporation}
\makebox[0.30\textwidth][c]{\textsuperscript{3}The University of Tokyo}
}

\input{def}

\newtheorem{dfn}{Definition}
\newtheorem{prop}{Proposition}

\newtheorem{formalthm}{Theorem}

\newtheorem{formalProposition}{Proposition}

\newtheorem{asm}{Assumption}
\usepackage{cancel}

\begin{document}

\maketitle
\begingroup
\renewcommand{\thefootnote}{}
\footnotetext{\textsuperscript{$\dagger$}Corresponding authors.}
\addtocounter{footnote}{-1}
\endgroup
\lhead{Preprint.}

\begin{abstract}
Uniform discrete diffusion models (UDMs) commonly use explicit time conditioning, but we find that it can often be unnecessary in practice. In this paper, we first show that the population-optimal UDM predictor generally depends on time: time controls how much the model should trust the observed context. We then show that this dependence can become negligible in finite-data settings relevant to language. When a corrupted training sequence remains much closer to its original clean sequence than to competing training sequences, the empirical-optimal predictor is nearly insensitive to time over most of the diffusion trajectory, where the guarantee weakens toward the high-noise endpoint. Empirically, trained language UDMs exhibit limited time sensitivity over most of the trajectory, while time-agnostic predictors remain competitive with, and often outperform, time-conditioned models across datasets and training objectives. These results challenge the use of explicit time conditioning in UDMs: although the population optimum depends on time, explicitly conditioning on it may often be unnecessary in practice.
\end{abstract}

\section{Introduction}
Diffusion models usually tell the neural network how noisy its input is~\citep{ho2020denoising,song2021score,lai2025principles}.
This information, commonly provided through \emph{time conditioning}, allows the same model to make different predictions at different stages of denoising.
But does the model actually need to be told the time? Recent work on continuous diffusion suggests that explicit time conditioning may not always be essential.
\citet{sun2025noise} show that removing time conditioning causes only a small performance degradation in several settings.
Separately, \citet{wang2025equilibriummatchinggenerativemodeling} develop a generative formulation whose neural model is time-independent, further showing that generation need not always be built around an explicitly time-conditioned predictor.

For discrete diffusion models, however, the answer depends strongly on the corruption mechanism.
In masked diffusion models (MDMs), corrupted tokens are replaced by a special \texttt{[MASK]} symbol.
The corrupted sequence therefore reveals which positions have been corrupted, and the population-optimal clean-token predictor does not need the diffusion time explicitly~\citep{zheng2024maskeddiffusionmodelssecretly}.
Consistent with this property, time-agnostic MDMs achieve comparable empirical performance~\citep{sahoo2024simple} and have been adopted in large language diffusion models~\citep{nie2025llada,ye2025dream}.

Uniform diffusion models (UDMs) are fundamentally different.
A corrupted token is replaced by an ordinary vocabulary token sampled from the uniform distribution, so the model cannot directly tell whether an observed token was preserved from the clean sequence or produced by noise~\citep{liu2025thinkgeneratediscretediffusion,amin2025maskingdiffusionworkscondition}.
The diffusion time can therefore carry information that is absent from the corrupted sequence itself, by indicating how much the observed context should be trusted.
This makes it unclear whether the time-agnostic property of MDMs can extend to UDMs.

This question has become increasingly important as UDMs scale to large language modeling.
Recent studies show that scaled UDMs can outperform MDMs in several settings~\citep{sahoo2026scalingmaskeddiffusionlanguage,vonrutte2026scalingbehaviordiscretediffusion}, followed by the development of large language UDMs~\citep{googledeepmind2026diffusiongemma,sahoo2026uno}.
Some of these models already use time-agnostic predictors, yet it remains unclear why this works, when time actually changes the optimal prediction, and how much explicit time conditioning helps in practice.

Throughout this work, we call a UDM \emph{time-agnostic} when its neural predictor does not receive diffusion time as an explicit input.
The forward corruption process, noise schedule, and other analytic time-dependent quantities remain unchanged.
We ask:
\begin{adjustwidth}{1.5em}{1.5em}
    \textit{What role does time play in UDMs, and when do UDMs need explicit time conditioning?}
\end{adjustwidth}
\noindent\textbf{First, we characterize why and how the population-optimal UDM predictor depends on time.}
Uniform corruption creates uncertainty about which observed tokens are reliable. Consequently, every subset of observed tokens constitutes a possible hypothesis about the reliable context. We show that time only controls the relative weights assigned to these hypotheses.
Thus, unlike MDMs where corrupted positions are explicitly identified, the population-optimal UDM predictor is generally time-dependent, but its dependence on time has a simple and structured form.

\noindent\textbf{Second, we show that finite training data can strongly suppress this dependence.}
Population-level time dependence arises from reweighting competing hypotheses about which observed tokens are reliable. In finite language data, however, the enormous discrete sequence space leaves training examples highly separated. A corrupted sequence therefore remains much closer to its originating clean sequence than to competing sequences, causing the posterior to collapse onto few hypotheses and weaken time-dependent reweighting. Only near the high-noise endpoint, where multiple hypotheses become plausible, can time affect the prediction. We make this intuition quantitative, deriving a bound that links the time sensitivity of the empirical oracle to the post-corruption separation margin. We empirically verify that language data exhibit large separation over most noise levels.

\noindent\textbf{Finally, we find that learned language UDMs make limited use of explicit time over most of the diffusion trajectory.}
The above finite-data result concerns the empirical oracle, not a learned neural network.
A learned model may recover additional time dependence through generalization or architectural inductive bias; we therefore separately analyze trained UDMs.
Their measured time sensitivity is weak over much of the trajectory and becomes substantially larger in the high-noise regime.
The predictive benefit of time conditioning is similarly concentrated near high noise.
A hybrid model that enables time conditioning only in this regime performs nearly the same as a fully time-conditioned model, while fully time-agnostic UDMs remain competitive with, and in several settings outperform, their time-conditioned counterparts across datasets and training objectives.

In summary, our results shift the question from whether time matters in UDMs to \emph{where it matters enough to justify explicit conditioning}.
This helps explain why time-agnostic UDMs can work well despite a time-dependent population optimum, and suggests a broader design space for simpler architectures, training methods, and samplers that use time only where it is useful.

\section{Background}
\label{sec:background}
We briefly introduce the UDM formulation and the recent variants used in this work.
See Appendix~\ref{sec:related_works} for more details and related work.

\noindent\textbf{Notation.}
Let $\calX=\{\ve_1,\dots,\ve_K\}\subset\Delta_K$ be a finite token space, where
each token is represented
as a one-hot vector and $\Delta_K$ is the $K$-category probability simplex. The vocabulary size is $K=|\calX|$, and the uniform prior is
$K^{-1}\vone_K$.
A sequence is $\vx=(\vx^{1},\ldots,\vx^{L})\in\calX^L$, and $\vz_t$ denotes
its corrupted version at time $t\in[0,1]$, where $t=0$ is data and $t=1$ is the
uniform prior.
$\alpha_t$ denotes a decreasing noise schedule with
$\alpha_0\approx1$ and $\alpha_1\approx0$, and 
$\rho_t\coloneqq K\alpha_t/(1-\alpha_t)$. For simplicity, we use the standard $\alpha_t=1-t$ hereafter; formal theoretical results for general schedules are provided in Appendix where the distinction matters.
Unless stated otherwise, expectations are over
$\vx\sim p_{\mathrm{data}}$ and $\vz_t\sim q_t(\cdot\mid\vx)$.
$\langle\cdot,\cdot\rangle$, $\odot$, and $\ind\{\cdot\}$ denote the dot product, Hadamard product, and indicator function, respectively.

\noindent\textbf{Uniform diffusion models.}
A UDM independently corrupts each token toward the uniform prior:
\begin{align}
q_t(\vz_t^{i}\mid\vx^{i})
=
\mathrm{Cat}\left(
\vz_t^{i};
\alpha_t\vx^{i}+(1-\alpha_t)K^{-1}\vone_K
\right).\label{eq:forward}
\end{align}
For $s<t$, the single token posterior $q_{s|0,t}$ follows from Bayes' rule using the $s\to t$ transition and the $s$-marginal (\eqref{eq:appendix_posterior}).
\citet{schiff2024simple} parameterize the reverse process by plugging $\vx_\theta^{i}(\vz_t,t)$ into this posterior,
\[
p_{s|t}^\theta(\vz_s^{i}\mid\vz_t)
\coloneqq
q_{s|0,t}
\left(
\vz_s^{i}
\mid
\vx^{i}=\vx_\theta^{i}(\vz_t,t),
\vz_t
\right),
\]
as detailed in \eqref{eq:reverse_process}.
The standard UDM is trained with the continuous-time NELBO:
\begin{align}
\label{eq:udm-elbo}
\Ls_{\mathrm{UDM}}(\theta)
=\int_0^1\E_{\vx,\vz_t}\sum_{i=1}^L
\frac{-\alpha_t'}{K\alpha_t}
\left[
\frac{K}{\bar{x}_{\theta,\ell_i}^{i}}
-\frac{K}{\bar{x}_{\ell_i}^{i}}
+\sum_{j\neq \ell_i}
\frac{\bar{x}_{j}^{i}}{\bar{x}_{\ell_i}^{i}}
\log\frac{\bar{x}_{\theta,\ell_i}^{i}\bar{x}_{j}^{i}}
{\bar{x}_{\theta,j}^{i}\bar{x}_{\ell_i}^{i}}
\right]\mathrm{d}t ,
\end{align}
where $\bar \vnu^{i}\coloneqq K\alpha_t\vnu^{i}+(1-\alpha_t)\vone_K$ and
$\ell_i$ denotes the non-zero index of $\vz_t^{i}$.

\noindent\textbf{Recent UDM formulations.}
Among several recent UDM formulations, we focus on two representative frameworks:
Diffusion Duality~\citep{sahoo2025diffusionduality} and the leave-one-out formulation of
\citet{gourevitch2026uniformdiffusionmodelsrevisited}.
Diffusion Duality retains the same uniform forward process and UDM NELBO, but represents the uniform-state marginal through the $\argmax$ of a Gaussian corruption and uses a curriculum from continuous relaxations to discrete inputs.
In contrast, \citet{gourevitch2026uniformdiffusionmodelsrevisited} show that
the neural predictor in the UDM bridge has the \emph{leave-one-out} (LOO)
posterior as its population optimum: each clean token is predicted from the
corrupted context excluding the observation at that same position, see
\eqref{eq:intrinsic_cavity_target}.
They further derive an analytic local-token correction that converts the LOO
predictor into the full denoising posterior, allowing the underlying predictor
to be trained with a simple cross-entropy loss.
We refer to this CE-trained formulation as \textsc{LOO+CE}.

\section{The Role of Time at the Population Optimum in UDMs}\label{sec:time_agnostic_udm}

The corruption mechanism determines whether
time needs to enter the clean-token predictor.
In MDMs, the special mask symbol reveals which positions have been corrupted,
so the population-optimal predictor can rely directly on the visible tokens and
is time-agnostic~\citep{zheng2024maskeddiffusionmodelssecretly}.

UDMs are different.
Because uniform corruption replaces a token by another ordinary vocabulary
token, the model cannot directly tell whether an observed token was copied from
the clean sequence or produced by noise. Time can therefore be informative because it changes how likely the observed
context is to have been preserved from the clean sequence.
This motivates two questions:
\begin{enumerate}
    \item Is the population-optimal UDM predictor inherently time-dependent?
    \item If so, through what structure does time affect its prediction?
\end{enumerate}
The first determines whether removing explicit time conditioning changes the
population optimum; the second reveals what information time actually contributes.

\subsection{The Population Target of the Uniform Channel}
\label{sec:uniform_channel_target}

The UDM reverse kernel has a seemingly complicated plug-in form
~\citep{gourevitch2026uniformdiffusionmodelsrevisited,sahoo2025diffusionduality,schiff2024simple}:
\begin{align}
\textstyle
p_{s|t}^{\theta}\!\left(\vz_s^{i}\mid \vz_t\right)
=q_{s|0,t}(\vz_s^{i}\mid\vx^{i}=\vx_{\theta}^{i}(\vz_t,t),\vz_t)=
\mathrm{Cat}\!\left(
\vz_s^{i};
\frac{
\rho_t \vz_t^{i} \odot \vx_{\theta}^{i}(\cdot,t)
+\frac{\rho_t}{\rho_s}\vz_t^{i}+
\left(1-\frac{\rho_t}{\rho_s}\right)
\frac{\rho_s\vx_{\theta}^{i}(\cdot,t)+\vone_K}{K+\rho_s}
}{
\rho_t
\left\langle
\vz_t^{i},
\vx_{\theta}^{i}(\vz_t,t)
\right\rangle
+1
}
\right).\label{eq:reverse_process}
\end{align}
Importantly, all neural-model dependence enters through the clean-token
prediction $\vx_\theta(\vz_t,t)$.
Thus, rather than analyzing the reverse kernel directly, we ask what this
predictor should output at the population optimum and how that prediction
depends on time.

\citet{gourevitch2026uniformdiffusionmodelsrevisited} show that the neural
prediction used by the UDM bridge has a \emph{leave-one-out} (LOO) posterior
as its population target.
Fix a target position $i\in[L]$.
Let $X\sim p_{\mathrm{data}}$ denote a random clean sequence,
$T\sim\operatorname{Unif}([0,1])$ the corruption time, and
$Z\mid(X=\vx,T=t)\sim q_t(\cdot\mid\vx)$ its corrupted version.
We write $Z^{-i}$ for the corrupted sequence with position $i$ removed,
and analogously for $\vz^{-i}$.
Thus, $Z^{-i}=\vz^{-i}$ means that all corrupted tokens except the target
position are observed.

The LOO target predicts the clean token at position $i$ from this remaining
corrupted context:
\begin{align}
    \vx_{\star}^{i}(\vz,t)
    \coloneqq
    \E
    \left[
        X^{i}
        \,\middle|\,
        Z^{-i}=\vz^{-i},T=t
    \right].
    \label{eq:intrinsic_cavity_target}
\end{align}
Since $X^i$ is one-hot, $\vx_{\star,k}^{i}(\vz,t)$ is the probability that
the clean token is $\ve_k$ given the other corrupted tokens.
The local observation $\vz^i$ can then be incorporated exactly through the
known uniform corruption likelihood: $\Pr\!\left(
X^i=\ve_k\mid Z=\vz,T=t
\right)
\propto
q_t\!\left(\vz^i\mid X^i=\ve_k\right)
\vx_{\star,k}^{i}(\vz,t)$. That is, $\vx_{\star}^{i}$ summarizes what the surrounding corrupted
tokens imply about the clean token, while the bridge uses the known corruption
process to account for the token observed at position $i$ itself.

Importantly, this LOO posterior is exactly the population minimizer of the UDM
NELBO in \eqref{eq:udm-elbo}:
\begin{align*}
    \vx_{\theta^\star}(\vz,t)
    =
    \vx_{\star}(\vz,t),\quad\text{for }P_{T,Z}\text{-almost every } (t,\vz).
\end{align*}
Substituting $\vx_\star$ into \eqref{eq:reverse_process} gives the corresponding
population-optimal reverse kernel:
\begin{align*}
p_{s|t}^{\theta^\star}\!\left(\vz_s^{i}\mid \vz_t\right)=q_{s|0,t}(\vz_s^{i}\mid\vx^{i}=\vx_{\star}^{i}(\vz_t,t),\vz_t).
\end{align*}

\eqref{eq:intrinsic_cavity_target} tells us what the optimal predictor is, but it does not yet reveal \emph{how} time changes its prediction: time appears only inside the conditioning distribution. To make this dependence explicit, recall that each token under uniform corruption can be viewed as arising either from the copy branch or from the uniform-noise branch. We therefore decompose the prediction over \textit{hypotheses} about which context positions were copied from the clean sequence, which leads to the following property.

\begin{restatable}[Explicit time decomposition of the population-optimal UDM predictor]{prop}{optimum}
\label{prop:optimum}
Fix $i\in[L]$ and a corrupted observation $\vz$.
For any index set $S\subseteq[L]\setminus\{i\}$, write
\[
\vz[S]
    \coloneqq
    \left(
        \vz^{j}
    \right)_{j\in S},\qquad X[S]
    \coloneqq
    \left(
        X^{j}
    \right)_{j\in S}
\] 
Then the population-optimal predictor satisfies
\begin{gather}
    \vx_{\star}^{i}(\vz,t)
    =
    \sum_{S\subseteq [L]\setminus\{i\}}
    \Pr(S\mid Z^{-i}=\vz^{-i},T=t)
    \E
    \left[
        X^{i}
        \,\middle|\,
        X[S]=\vz[S]
    \right],
    \label{eq:optimum}\\
    \Pr(S\mid Z^{-i}=\vz^{-i},T=t)
    \propto
    \rho_t^{|S|}
    \Pr\!\left(
        X[S]=\vz[S]
    \right).
    \nonumber
\end{gather}
Here,
$\Pr(X[S]=\vz[S])
=
\sum_{\vx:\,\vx[S]=\vz[S]}p_{\mathrm{data}}(\vx)$
is the data probability mass of the partial observation $\vz[S]$ under the
hypothesis index set $S$.
\end{restatable}
Proposition~\ref{prop:optimum} makes the role of time explicit: \textbf{time controls how much of the hypothesis $\vz[S]$ should be trusted}.
Given a corrupted context $\vz^{-i}$, the model does not know which observed
tokens were from the clean sequence or from
uniform noise.
It therefore considers every subset
$S\subseteq[L]\setminus\{i\}$ as a possible set of copied positions.
If the positions in $S$ were indeed copied, then
$\E\!\left[
    X^{i}
    \,\middle|\,
    X[S]=\vz[S]
\right]$
is the corresponding prediction of the target token from that clean context.
The final prediction averages over all such possibilities according to their
posterior weights. 

We now interpret this time-weighted posterior in more detail. Using $\rho_t=K\alpha_t/(1-\alpha_t)$, 
\begin{align*}
\Pr(S\mid Z^{-i}=\vz^{-i},T=t)
\propto
\rho_t^{|S|}
\Pr\!\left(X[S]=\vz[S]\right)
=
\underbrace{
\Pr\!\left(X[S]=\vz[S]\right)
}_{\substack{\text{Data}\\\text{dependent}}}
\underbrace{
K^{|S|}
}_{\substack{\text{Fixed vocabulary}\\\text{factor}}}
\underbrace{
\left(\tfrac{\alpha_t}{1-\alpha_t}\right)^{|S|}
}_{\substack{\text{Time}\\\text{dependent}}}.
\end{align*}
This factorization cleanly separates the sources of the posterior weighting.
The data-dependent term measures how compatible the observed values on $S$ are
with the data distribution.
The time-dependent term changes the relative preference for subsets of
different sizes as the corruption level varies, while $K^{|S|}$ is a fixed
channel factor that amplifies this size dependence.

Near the clean endpoint, $\rho_t$ is large, so the model places more weight on
explanations in which many observed tokens were copied from the clean sequence.
As noise increases, $\rho_t$ decreases and explanations with fewer copied
positions receive relatively more weight.
Thus, unlike in MDMs, the population-optimal UDM predictor is generally
time-dependent: $\vx_{\star}^{i}(\vz,\cdot)\not\equiv \mathrm{const}$.

\subsection{Time Acts Through the Amount of Matched Context}
\label{sec:reliability_geometry}

Proposition~\ref{prop:optimum} expresses the optimal prediction through all
$2^{L-1}$ possible hypotheses of copied context positions.
We can further simplify this optimum using the symmetry of the uniform channel
(\eqref{eq:forward}): every position at which a candidate clean sequence
matches the corrupted observation contributes the same factor $1+\rho_t$.
Hence, the likelihood depends on a matching set $M \subseteq [L]$ only through
its size, via $(1+\rho_t)^{|M|}$, regardless of the token identities, the locations of the matches, or whether a matching token was copied from clean sequence or independently produced by uniform corruption. Thus, time acts only through \emph{how many} context positions match.
This motivates the following statistic.

\begin{dfn}[Match count random variable]
For random variables $X$ and $Z$, and a fixed target position $i$, define
\[
    M^i(X,Z)
    \coloneqq
    \sum_{j\in[L]\setminus\{i\}}
    \ind\{X^{j}=Z^{j}\}.
\]
\end{dfn}
Thus, $M^i$ counts how many context tokens of a candidate clean sequence agree
with the corrupted observation, excluding the target position $i$, without recording which particular positions match.

\begin{restatable}[Time dependence through the matched-context count]{thm}{maskcount}
\label{thm:exponential_reliability}
For every target position $i$,
\[
    X^{i}
    \perp
    T
    \mid
    Z^{-i},M^i,
\]
i.e., $X^i$ and $T$ are conditionally independent given $Z^{-i}$ and $M^i$. Moreover, \eqref{eq:intrinsic_cavity_target} admits 
the exact decomposition
\begin{gather*}
    \vx_{\star}^{i}(\vz,t)
    =\sum_{m=0}^{L-1}w_m^i(\vz,t)\mathbf h_m^{i}(\vz),
    \label{eq:exponential_reliability_mixture}\\
    w_m^i(\vz,t)\coloneqq
    \Pr
    \left(
        M^i=m
        \,\middle|\,
        Z^{-i}=\vz^{-i},
        T=t
    \right)
    \propto
    (1+\rho_t)^m
    \Pr
    \left(
        M^i(X,\vz)=m
    \right),\\
    \mathbf h_m^{i}(\vz)
    \coloneqq
    \E
    \left[
        X^{i}
        \,\middle|\,
        M^i=m,Z^{-i}=\vz^{-i}
    \right],
\end{gather*}
where
$\Pr(M^i(X,\vz)=m)
=\sum_{M^i(\vx,\vz)=m}p_{\mathrm{data}}(\vx)$
is the data probability mass of clean sequences that match $\vz$ in exactly
$m$ context positions.
\end{restatable}

Theorem~\ref{thm:exponential_reliability} gives a compact description of the
entire time dependence.
Once the corrupted context and its match count $M^i$ are fixed, the
distribution of the clean token $X^i$ no longer depends on time.
The component prediction $\mathbf h_m^i(\vz)$ is therefore time-independent;
time only changes how strongly different match counts $m$ are weighted through
$(1+\rho_t)^m$.
In other words, the population-optimal UDM does not learn an unrelated
clean-token prediction at every noise level.
Time acts through a single scalar $\rho_t$ that reweights predictions
associated with different amounts of matched context.

\subsection{Characterizing the Influence of Time in Uniform Diffusion}
\label{sec:temporal_sensitivity}

Theorem~\ref{thm:exponential_reliability} shows \emph{where} time enters the
population-optimal prediction: it changes the weights over the match count
$M^i$.
We next quantify \emph{how strongly} this reweighting changes the prediction.

\begin{restatable}[Exact temporal sensitivity]{prop}{derivative}
\label{prop:exact_temporal_sensitivity}
Let $\eta_t=\log(1+\rho_t)$ and let $\ve_k$ denote the $k$-th one-hot vector.
Then the temporal derivative of the context posterior is
\begin{align*}
    \frac{\partial}{\partial\eta_t}
    \vx_{\star}^{i}(\vz,t)
    =
    \operatorname{Cov}_{t,\vz}
    \left(
        \mathbf h_{M^i}^{i}(\vz),
        M^i
    \right),
\end{align*}
where the covariance is taken componentwise.
Equivalently, for any candidate category $k$ with $\vx_{\star,k}^{i}(\vz,t)>0$,
\begin{align}
    \frac{\partial}{\partial\eta_t}
    \log
    \vx_{\star,k}^{i}(\vz,t)
    =
    \E_{t,\vz}
    \left[
        M^i
        \,\middle|\,
        X^{i}=\ve_k
    \right]
    -
    \E_{t,\vz}[M^i],
    \label{eq:single_candidate_reliability_sensitivity}
\end{align}
where the subscript $(t,\vz)$ denotes the posterior distribution
$\Pr(\cdot\mid Z^{-i}=\vz^{-i},T=t)$.
\end{restatable}

Proposition~\ref{prop:exact_temporal_sensitivity} makes the effect of time
particularly transparent.
Larger $\eta_t$ corresponds to a cleaner channel.
\eqref{eq:single_candidate_reliability_sensitivity} therefore says
that a candidate token $\ve_k$ gains probability toward the clean endpoint
when clean sequences predicting $\ve_k$ have more average matches
with the observed context than the posterior average.
If they have fewer matches, its probability decreases.
Thus, the strength of time dependence is governed by differences in matched-context support across candidate tokens.

\begin{takeawaybox}
\textbf{Population-level takeaway.}
The population-optimal UDM predictor is generally time-dependent, but time only reweights predictions across different match counts.
Its effect is strongest when competing clean candidates are
supported by different amounts of matched context.
\end{takeawaybox}

\section{Time Sensitivity of the Finite-Data Oracle}
\label{sec:analysis}

At the population level, a rich data distribution can assign substantial
probability to multiple clean sequences that plausibly explain the same
corrupted context.
In this regime, changing $t$ can meaningfully reweight these competing
explanations and thereby change the optimal prediction.

The situation can be very different under finite training data.
An empirical distribution places probability mass on only finitely many
sequences in the enormous discrete space $\calX^L$.
If these sequences are sufficiently separated, a corrupted training sample
$\vz$ can remain much more compatible with its originating sequence $\vx$
than with any competing sequence that predicts a different token at position
$i$.
The posterior may then already be strongly concentrated, leaving little room
for the time-dependent reweighting to change the resulting prediction.
This motivates the central question:

\begin{adjustwidth}{1.5em}{1.5em}
\emph{When does finite training data suppress the time sensitivity of the oracle UDM predictor?}
\end{adjustwidth}

We show that the answer is governed by the geometry of the finite training set: how well separated training sequences are in the enormous discrete sequence space. Our theoretical analysis proceeds in two steps: \textbf{1)} when training sequences are sufficiently far apart, the originating sequence is likely to remain separated from competing sequences after corruption, and \textbf{2)} whenever this separation is sufficiently large, the oracle Fisher sensitivity for corrupted sequence is strongly suppressed over a broad range of diffusion times, with the guarantee weakening toward the high-noise endpoint.

\subsection{When Finite High-Dimensional Data Suppress Time Sensitivity}
\label{sec:intuition}
We now make the finite-data mechanism precise.
Given $N$ training sequences $\vx_1,\ldots,\vx_N\in\calX^L$, define their empirical distribution as $\real
\coloneqq
\frac{1}{N}\sum_{n=1}^N\delta_{\vx_n}$. This is the finite-data counterpart of $p_{\mathrm{data}}$ considered in the previous section, with the oracle supported only on the observed training sequences.

\textbf{The first key question} is how much of the separation between distant training sequences survives even after corruption.
Consider a corrupted $\vz$ generated from a training sequence $\vx$.
For prediction at position $i$, only training sequences with $\vx'^i\neq\vx^i$ can support a different target token.
What matters is therefore whether the originating sequence $\vx$ retains a
match-count advantage over every such competitor after corruption.
We quantify this advantage through the following separation margin.

\begin{dfn}[Hamming separation and $\Delta$-separability]
\label{def:delta}
For $\vx\in\operatorname{supp}(\real)$, define its minimum Hamming distance
to the remaining training sequences as
\[
d(\vx)
\coloneqq
\min_{\vx'\in\operatorname{supp}(\real)\setminus\{\vx\}}
\sum_{j=1}^{L}
\ind\!\left\{\vx'^j\neq\vx^j\right\}.
\]
For an integer $\Delta\ge1$, we say that a corrupted sequence $\vz$ is
\emph{$\Delta$-separable with respect to $\vx$ at coordinate $i$} if
\begin{align}
M^i(\vx,\vz)-M^i(\vx',\vz)
\ge \Delta
\qquad
\forall\,
\vx'\in\operatorname{supp}(\real)
\text{ with }
\vx'^i\neq\vx^i.
\label{eq:delta_seperable}
\end{align}
\end{dfn}

Thus, $\Delta$ is exactly the match-count advantage retained by the
originating sequence over every training sequence that would predict a
different token at position $i$.
The following result shows that such an advantage survives when the training sequences are sufficiently separated.

\begin{prop}[$\Delta$-separation under UDM corruption (informal)]
\label{lem:delta-seperable}
Given a training sequence $\vx$, let
$T\sim\operatorname{Unif}([0,1])$ and
$Z\sim q_T(\cdot\mid\vx)$.
For any integer $\Delta\ge1$ and $i\in[L]$, when $K\gg L$ and $d(\vx)$ is
large,
\[
\Pr\!\left(
Z
\text{ is $\Delta$-separable w.r.t.\ $\vx$ at $i$}
\right)
\gtrsim
1-
\frac{\log N+\Delta}
{(1-e^{-1})d(\vx)}.
\]
\end{prop}

This simplified bound exposes the tradeoff directly.
If training sequences are sufficiently separated, \ie Hamming separation $d(\vx)$ is large, the source $\vx$ remains
distinguishable after corruption. Conversely, more competing sequences or a larger required margin $\Delta$ make the event harder.
Importantly, the dependence on dataset size is only logarithmic in $N$.
If $d(\vx)$ is too small relative to $\log N+\Delta$, the bound can become
uninformative.
The formal finite-$K$ statement is in Appendix~\ref{sec:formal_delta}.

\textbf{The second key question} is how separation suppresses the time sensitivity of the oracle prediction.
For fixed $\vz$, coordinate $i$, and time $t$, define the Fisher time
sensitivity
\begin{equation}
\mathcal I_t^{i}(\vz)
\coloneqq
\sum_{k:\,\vx_{\star,k}^{i}(\vz,t)>0}
\vx_{\star,k}^{i}(\vz,t)
\left(
\frac{\partial}{\partial t}
\log \vx_{\star,k}^{i}(\vz,t)
\right)^2
=
\sum_{k:\,\vx_{\star,k}^{i}(\vz,t)>0}
\frac{
\left(
\frac{\partial}{\partial t}
\vx_{\star,k}^{i}(\vz,t)
\right)^2
}{
\vx_{\star,k}^{i}(\vz,t)
}.
\label{eq:fisher_sensitivity}
\end{equation}
It measures how quickly the optimal categorical prediction changes when only
the diffusion time is varied.
Indeed, for infinitesimal $\delta$, we have $\mathcal{D}_{\mathrm{KL}}
\left[
\vx_\star^{i}(\vz,t)
\,\middle\|\,
\vx_\star^{i}(\vz,t+\delta)
\right]
=
\frac{\delta^2}{2}\mathcal I_t^{i}(\vz)
+
o(\delta^2)$. Hence, small $\mathcal I_t^{i}(\vz)$ means that nearby times produce nearly
indistinguishable oracle predictions.

The next result provides the key link: a large match-count advantage directly
limits this time sensitivity.

\begin{restatable}[Fisher sensitivity with finite data (informal)]{thm}{finitefisher}
\label{thm:finitefisher}
For $\Delta$-separable $\vz$ with respect to $\vx$ at coordinate $i$,
\[
\mathcal I_t^{i}(\vz)
\le
NL^2
\frac{
K^2 t^{\Delta-2}
}{
\bigl(K-(K-1)t\bigr)^{\Delta+2}
}.
\]
Moreover, for $\Delta\ge2$, the right-hand side is strictly increasing in
$t\in(0,1)$.
\end{restatable}
The bound becomes weaker toward the high-noise endpoint, where corruption makes
the originating sequence harder to distinguish from competing sequences.
At lower noise levels, a larger separation margin $\Delta$ keeps these
competitors much less plausible, leaving less room for changes in time to alter
the oracle prediction. If $\Delta\ge2$, a convenient consequence is:
\begin{equation}
\sup_{0<t\le1-K^{-1/4}}
\mathcal I_t^{i}(\vz)
\le
NL^2K^{-(3\Delta-2)/4}.
\label{eq:fisher_upper_loose_bound}
\end{equation}
For practical LLM vocabulary sizes, the interval
$(0,1-K^{-1/4}]$ already covers most of the diffusion trajectory.
More importantly, \eqref{eq:fisher_upper_loose_bound} can be inverted to ask
how much separation is sufficient for a desired sensitivity level.
To guarantee
$\sup_{0<t\le1-K^{-1/4}}\mathcal I_t^i(\vz)\le\zeta$, it is sufficient that $\Delta
\ge
\max\left\{
2,\,
\frac{2}{3}
+
\frac{4}{3}
\log_K
\frac{NL^2}{\zeta}
\right\}$. Thus, although the Fisher bound contains $N$ and $L^2$, the required
match-count advantage grows only logarithmically with dataset size and sequence
length. Each additional unit of $\Delta$ further suppresses the bound by a
factor of $K^{-3/4}$.

\paragraph{The bounds remain meaningful at LLM scale.}
Consider a K2-Horizon-375B-scale pretraining configuration~\citep{k2horizon2026} with
$N\approx1.9\times10^9$, $L=8192$, and $K=250{,}624$. Here, by \eqref{eq:fisher_upper_loose_bound}, only an integer margin $\Delta\ge7$ guarantees $\sup_{0<t\le0.955}\mathcal I_t^i(\vz)\le10^{-5}$.
In other words, even with billions of training sequences, the originating
sequence only needs to retain a seven-token match-count advantage over every
competitor predicting a different target token.
By the formal bound in Proposition~\ref{lem:delta-seperable} (Appendix~\ref{sec:formal_delta}), $d(\vx)\ge900$, about $11\%$ of the sequence length, already gives a
$0.95$ lower bound on $7$-separability, while $d(\vx)\ge451$ gives a
$0.9$.
Finally, small Fisher sensitivity controls finite changes in the oracle
prediction.
If
$\sup_{0<u\le1-\epsilon}\mathcal I_u^{i}(\vz)\le\zeta$, then for any
$t,s\in(0,1-\epsilon)$,
\[
\textstyle
\left\|
\sqrt{\vx_\star^{i}(\vz,t)}
-
\sqrt{\vx_\star^{i}(\vz,s)}
\right\|_2
\le
\frac{|t-s|}{2}\sqrt{\zeta}.
\]
Thus, once separation makes the Fisher bound small, the finite-data oracle is nearly invariant to time.

\subsection{Oracle Has Near-Zero Time Sensitivity on Real Language Data}
\label{sec:numerical_analysis}

\begin{figure}[t]
    \centering
    \includegraphics[width=\linewidth]{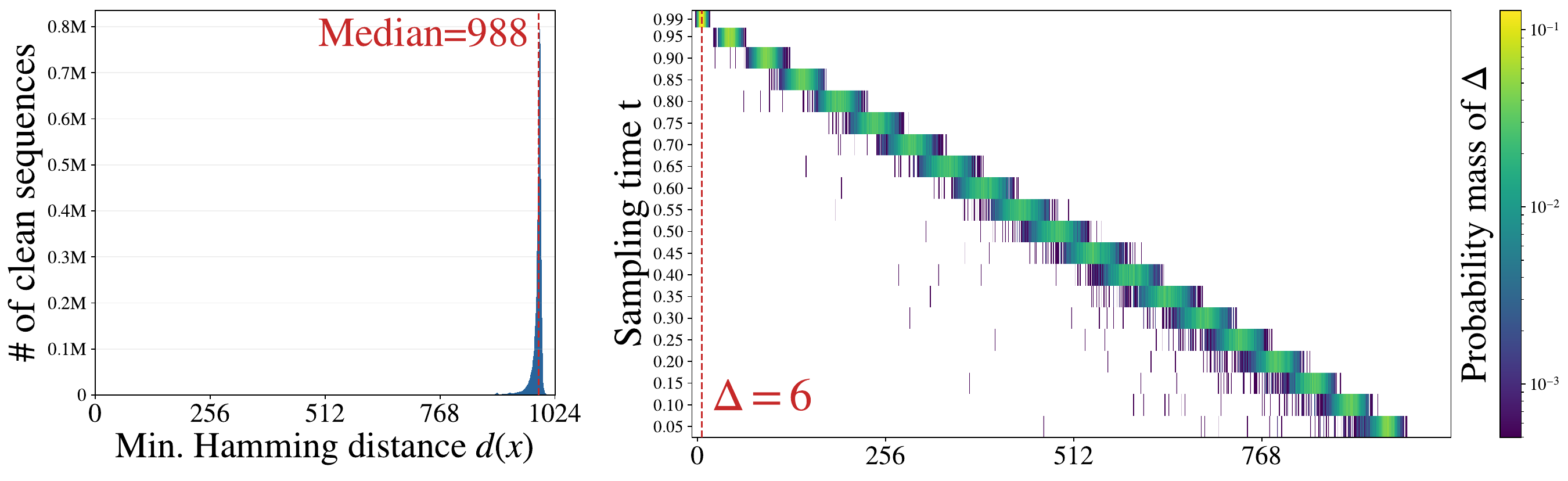}
    \vspace{-0.2cm}
    \caption{
    Empirical separation of OWT training data.
    \textit{Left}: distribution of the minimum Hamming distance $d(\vx)$ from
    each training sequence to the remaining data, with median $d(\vx)=988$.
    \textit{Right}: distribution of the largest post-corruption separation
    margin $\Delta$ across diffusion times.
    The dashed line marks $\Delta=6$, which implies an oracle Fisher-sensitivity
    bound below $10^{-5}$ for $t\le0.933$.
    Small separation occurs primarily when $\vz$ is sampled near the extreme high-noise endpoint.
    }
    \label{fig:data_separation}
    \vspace{-0.5cm}
\end{figure}

We now examine whether the finite-data regime predicted by
Proposition~\ref{lem:delta-seperable} and
\eqref{eq:fisher_upper_loose_bound} \textit{actually occurs} in real language data.
We consider OWT~\citep{Gokaslan2019owt}, a standard benchmark in discrete diffusion language
modeling, with $K=50{,}258$,  $L=1024$, and $N=8.7\times10^6$. 
For these values,
$\sup_{0<t\le0.933}\mathcal I_t^i(\vz)\le\zeta$
with $\zeta=10^{-5}$ only requires $\Delta\ge6$.
The formal bound in Appendix~\ref{sec:formal_delta} further shows that
$d(\vx)\ge698$ is sufficient for at least a $0.95$ lower bound on
$6$-separability.

We now compute $d(\vx)$ and the post-corruption separation margin $\Delta$ directly on OWT. Before diffusion corruption, clean OWT is highly separated in Hamming space.
As shown in Figure~\ref{fig:data_separation} (\textit{Left}), the distribution
of $d(\vx)$ is concentrated near the maximum sequence length $L=1024$, with
median $d(\vx)=988$, well above the sufficient value $698$. We next test whether this separation survives UDM corruption.
For each $t\in\{0.05,0.1,\ldots,0.95,0.99\}$, we sample $2000$ pairs
$\vx\sim\real$ and $\vz_t\sim q_t(\cdot\mid\vx)$ and measure the largest
$\Delta$ for which $\vz_t$ is $\Delta$-separable.
Figure~\ref{fig:data_separation} (\textit{Right}) shows that the observed
margin remains above $6$ over almost the entire diffusion trajectory, becoming small primarily when $\vz$ is sampled near the extreme high-noise endpoint.

These measurements confirm that OWT lies in the separation regime predicted by our theory: most of the training pairs $(\vx, \vz)$ satisfies 6-separability, and whenever the observed margin satisfies $\Delta\ge6$,
\eqref{eq:fisher_upper_loose_bound} gives
$\sup_{0<t\le0.933}\mathcal I_t^i(\vz)\le10^{-5}$,
implying negligible oracle time sensitivity over most of the diffusion
trajectory.
Near the extreme high-noise endpoint, smaller separation margins weaken this
guarantee and leave more room for time to influence the oracle prediction.
\begin{takeawaybox}
\textbf{Finite-data takeaway.}
When finite training sequences remain well separated after corruption, the
originating sequence retains a clear match-count advantage and the oracle
becomes nearly insensitive to time over most of the diffusion trajectory.
This regime is empirically observed in practical language data, with the separation weakening mainly near the high-noise
endpoint.
\end{takeawaybox}

\section{You Might Not Need Time in UDMs for Text Generation}
\label{sec:results}

Sections~\ref{sec:time_agnostic_udm} and~\ref{sec:analysis} reveal a tension:
the population-optimal UDM predictor generally depends on time, while the
finite-data oracle can be nearly time-insensitive over most of the time.
The remaining question is whether a learned UDM actually benefits from
time conditioning.
Two possibilities arise:
\vspace{-0.22cm}
\begin{itemize}
    \item \textbf{Time-variant UDMs may benefit from recovering useful time dependence} through generalization or inductive bias beyond the finite-data oracle.

    \item \textbf{Time-agnostic UDMs may benefit from removing largely unnecessary time dependence.}
    If time dependence remains weak over most of the trajectory, removing explicit time conditioning may simplify the model and improve learning.
\end{itemize}
\vspace{-0.2cm}
We test these directly by measuring the time sensitivity of
learned UDMs and comparing time-conditioned and time-agnostic models.
Throughout this section, \textsc{Duality} denotes the framework of
\citet{sahoo2025diffusionduality}, while \textsc{LOO+CE} denotes the
LOO denoiser trained with cross-entropy
\citep{gourevitch2026uniformdiffusionmodelsrevisited}.
See Appendix~\ref{sec:detailed_background} for framework details and
\ref{sec:exp_settings} for experimental settings.

\subsection{Learned UDMs Exhibit Limited Time Dependence in Language}
\label{sec:language}

\begin{figure}[t]
    \centering
    \includegraphics[width=\linewidth]{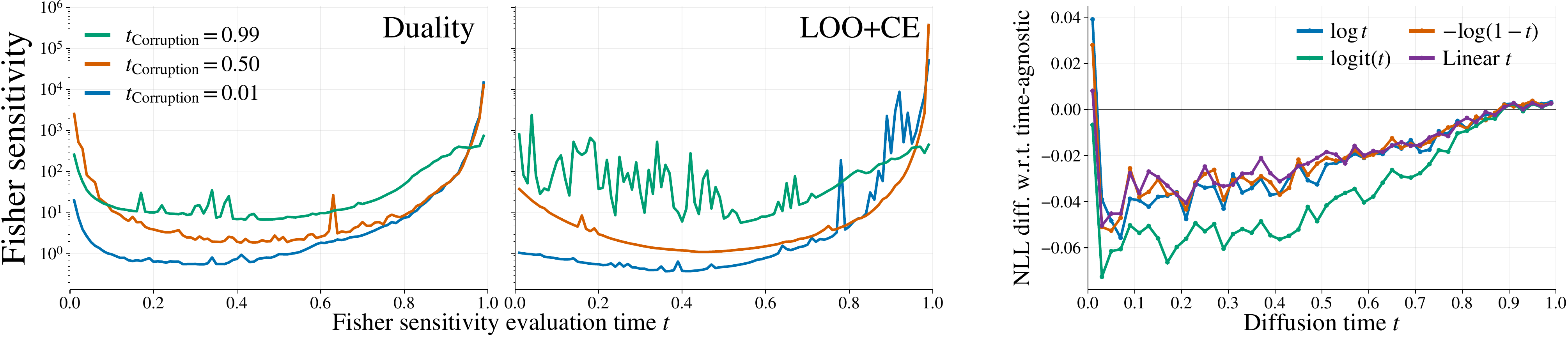}
    \caption{
        \textit{Left}: Fisher sensitivity of \textsc{Duality} and \textsc{LOO+CE}, varying $\vz$ sampling time $t_{\mathrm{corruption}}\in\{0.01,0.50,0.99\}$ and evaluating Fisher sensitivity on every time coordinate.
        \textit{Right}: Per-time NLL difference between time-agnostic and
        time-variant UDMs. 
    }
    \label{fig:learned_time_dependence}
\end{figure}

\noindent\textbf{Learned time sensitivity follows the finite-data prediction.}
We first examine whether the time sensitivity of learned UDMs exhibits the qualitative behavior predicted by our finite-data analysis.
Using the official OWT checkpoints of \textsc{Duality} and \textsc{LOO+CE}, we sample corrupted sequences $\vz$ at
$t_{\mathrm{corruption}}\in\{0.01,0.50,0.99\}$.
For each fixed $\vz$, we then vary the time input used for prediction and measure its Fisher sensitivity across the entire diffusion trajectory.

As shown in Figure~\ref{fig:learned_time_dependence} (\textit{left}),
the learned models qualitatively match our finite-data analysis:
First, Fisher sensitivity increases with $t_{\mathrm{corruption}}$.
Stronger corruption tends to reduce the separation margin $\Delta$, weakening the finite-data suppression guarantee in \eqref{eq:fisher_upper_loose_bound}.
Second, for a fixed corrupted input, Fisher sensitivity generally increases with the evaluation time and rises sharply near the high-noise endpoint, where the bound in Theorem~\ref{thm:finitefisher} becomes weaker.
Finally, we use held-out test sequences for this analysis, although our finite-data theory is stated for the empirical training distribution.
The similar qualitative behavior on held-out data suggests that limited time sensitivity can extend beyond training examples, while our theory does not directly establish this behavior off the empirical support.

\noindent\textbf{The localized time sensitivity is robust and provides limited practical benefit.}
We next ask whether this concentration of time sensitivity can be resolved by how time is parameterized to the model.
We train conventional UDMs~\citep{schiff2024simple} on LM1B.
We keep the forward process unchanged and reparameterize only the scalar passed to the time embedding as $\hat t=f(t)$, where
$f(t)\in\{-\log(1-t),\log t,\operatorname{logit}(t),t\}$.
These parameterizations allocate different resolution across diffusion times.
Nevertheless, Figure~\ref{fig:learned_time_dependence_appendix} in Appendix shows a consistent pattern across all parameterizations: learned UDMs exhibit weak time sensitivity except near the high-noise endpoint.

We next compare the time-conditioned model with its time-agnostic counterpart.
Figure~\ref{fig:learned_time_dependence} (\textit{Right}) shows that time conditioning improves per-time NLL mainly near $t=1$, precisely where the learned Fisher sensitivity becomes substantial.
Over most of the diffusion trajectory, however, the time-agnostic model achieves lower NLL and also performs better overall.
Together, these results suggest that learned UDMs make substantial use of explicit time conditioning only in a limited high-noise region, and that modeling this localized dependence provides limited practical benefit.

\subsection{Time-agnostic UDM versus Time-variant UDM}

We next compare time-agnostic and time-variant UDMs across broader settings, including different datasets, tokenizers, and training objectives.
Beyond the conventional UDM studied in Section~\ref{sec:language}, we now evaluate \textsc{Duality} and \textsc{LOO+CE}.
We follow widely adopted experimental settings in discrete diffusion language modeling~\citep{sahoo2025diffusionduality,arriola2025bd3lm,schiff2024simple}, evaluating models on LM1B~\citep{chelba2013lm1b}, both with and without sentence packing, and OWT, all trained for 1M steps, with the same backbone.  See Appendix~\ref{sec:exp_settings} for more detailed settings. 

\input{table}

\noindent\textbf{Results on LM1B.} We compare four setttings for time conditioning: 1) Time-variant model, 2) Time-agnostic model where we remove the time-modulation layers, reducing the parameter count from 139M to 131M, 3) Time-agnostic model with 139M parameters obtained by adding one Transformer block to 2), and 4) Hybrid model that only activates time-modulation layer when $t>0.8$. Table~\ref{tab:best-validation-ppl} shows that the hybrid model that only learns time-sensitivity on high-noise regime is almost equivalent to the time-variant model, and moreover, time-agnostic UDMs remain consistently competitive with their time-dependent counterparts across training frameworks.

\begin{wraptable}{r}{0.33\textwidth}
    \centering
    \vspace{-0.5cm}
    \caption{Validation PPL on OWT.}
    \vspace{-0.3cm}
    \label{tab:owt_validation_ppl}
    \small
    \begin{tabular}{lcc}
        \toprule
        & Agnostic & Variant \\
        \midrule
        \textsc{Duality}  & 24.33 & 25.54 \\
        \textsc{LOO+CE} & 25.29 & 25.29 \\
        \bottomrule
    \end{tabular}
    \vspace{-0.5cm}
\end{wraptable}
\noindent\textbf{Results on OWT.}
We compare time-variant and time-agnostic models with the same parameter size. As shown in Table~\ref{tab:owt_validation_ppl}, time-agnostic UDMs are comparable to time-variant UDMs. Generative PPL was comparable for \textsc{Duality}, while the time-agnostic model performed better for \textsc{LOO+CE}; see Appendix~\ref{sec:gen_ppl}.

\begin{takeawaybox}
\textbf{Learned-model takeaway.}
Time matters mainly near the high-noise endpoint. Consistently, time-agnostic UDMs remain competitive across datasets and objectives, suggesting that explicit time conditioning may offer limited benefit over most of the trajectory.
\end{takeawaybox}

\section{Conclusion}

We studied the role of time in population optimal UDMs, and why it weakens under finite training data. Consistently, learned UDMs exhibit limited time dependence over most of the trajectory, while time-agnostic models remain competitive across datasets and frameworks.
We expect this view opens a door to future research that allows a simpler foundation for architectures, training, and samplers.

\bibliography{main}
\bibliographystyle{main}

\clearpage
\appendix
\tableofcontents
\clearpage

\section{Background}\label{sec:related_works}
\subsection{Related Work}
\noindent\textbf{Discrete diffusion models.}
Diffusion generative modeling was originally introduced by \citet{sohl2015deep}, and has become a dominant approach for continuous domains such as images~\citep{ho2020denoising,song2021score}. By defining a time-variant stochastic differential equation~\citep{song2021score}, diffusion models establish a continuous path between the data and noise distributions. This success has motivated extensions to discrete domains,  leading to discrete diffusion models~\citep{Austin2021,Hoogeboom2021b,campbell2022continuous}. The forward corruption process is typically designed in one of two ways: (i) \emph{uniform} corruption, which replaces tokens with random vocabulary elements~\citep{lou2024discrete,sahoo2025diffusionduality}, or (ii) \emph{masking}-based corruption, which maps tokens to an absorbing \texttt{[MASK]} state~\citep{sahoo2024simple,shi2025simplified}. 

\noindent\textbf{Uniform discrete diffusion models and plug-in parametrization.}
Early uniform diffusion models (UDMs), together with masked diffusion models (MDMs), were formulated as discrete diffusion models by specifying particular forward corruption kernels~\citep{Austin2021,campbell2022continuous,lou2024discrete,Hoogeboom2021b}. 
Among those, the uniform bridge plug-in parameterization was first introduced in multinomial diffusion by \citet{Hoogeboom2021b}. Using uniform bridge plug-in parameterization, \citet{schiff2024simple} developed a continuous-time formulation and derived its NELBO, reporting improved performance over earlier uniform-diffusion baselines.
Building on this, \citet{sahoo2025diffusionduality} reinterpreted the uniform kernel as the $\argmax$ pushforward of a Gaussian corruption kernel and proposed a curriculum that initially uses a biased continuous-input relaxation before returning to discrete-input UDM training. More recently, \citet{gourevitch2026uniformdiffusionmodelsrevisited} characterized the unrestricted population optimum of the bridge plug-in predictor as a leave-one-out posterior, which enables training with a simple cross-entropy objective under a leave-one-out parametrization. 

Notably, recent studies have shown that UDMs can outperform MDMs when scaled up in the language domain~\citep{sahoo2026scalingmaskeddiffusionlanguage,vonrutte2026scalingbehaviordiscretediffusion}, and large UDMs such as DiffusionGemma~\citep{googledeepmind2026diffusiongemma} and Uno~\citep{sahoo2026uno} have recently been released.
Despite growing attention to the application of UDMs in the language domain, it remains unclear whether time conditioning is actually necessary for UDMs.
In this regard, our work takes the population-optimum characterization as a starting point and analyzes its dependence on time in detail, showing that, in the language domain, the time sensitivity of UDMs can be largely confined to the endpoint region of $t$.

\noindent\textbf{Time-agnostic masked diffusion models.}
Earlier continuous-time formulations for mask corruption kernels commonly used time-conditioned networks to parameterize the reverse process~\citep{campbell2022continuous,lou2024discrete}. However, \citet{zheng2024maskeddiffusionmodelssecretly,ou2024RADD} showed that MDMs do not inherently require time conditioning in the model, and \citet{sahoo2024simple} demonstrated empirically that time-agnostic MDMs can slightly outperform their time-variant counterparts. Consequently, recent large-scale MDMs~\citep{ye2025dream,nie2025llada} have largely adopted time-agnostic architectures. Consequently, time-agnostic MDMs provide a simpler foundation, leading to subsequent developments in inference~\citep{lee2025iterref} and reinforcement learning~\citep{hong2025improvingdiscretediffusionunmasking}. Accordingly, if time-agnostic modeling is also viable for UDMs, it would provide a simpler and more extensible framework. Unlike MDMs, however, where time-agnostic and time-variant models share the same population optimum, we show that the two optima generally differ for UDMs. Nevertheless, under a finite-data setting, we find that time-agnostic UDMs can still be practically viable.

\noindent\textbf{Time-agnostic continuous diffusion models.}
Recent studies have investigated when continuous diffusion models can dispense
with explicit time conditioning
\citep{sahraeeardakan2026geometrynoisediffusionmodels,
sun2025noise,helbling2026timeitdatageometry}.
In continuous diffusion, the state evolves in a differentiable continuous
space, and the reverse process is described locally by a score
$\nabla_{\vz}\log p_t(\vz)$ or, equivalently in the ODE view, by a vector
field over the continuous state space.
The current state can therefore encode \textit{where the sample lies along the
generative trajectory}, allowing even a time-independent vector field to evolve
according to its state.
This continuous dynamics of particles enables geometric explanations for time-agnosticity.
For example,
\citet{sahraeeardakan2026geometrynoisediffusionmodels} study a
time-independent vector field and relate the resulting dynamics to a marginal
energy landscape and gradient flow.
A complementary explanation is that the corrupted observation itself can reveal
the noise level.
\citet{sun2025noise} relate concentration of $T\mid Z=\vz$ to the gap between
time-agnostic and time-conditioned predictors, and further extend their analysis to
multiple data points under an i.i.d.\ Gaussian data model.
Similarly, \citet{helbling2026timeitdatageometry} construct a time estimator
$\hat t(\vz)$ under a spiked-covariance Gaussian model and show that the error
from hiding time becomes small in high dimensions.

In contrast, our analysis first reveals the distinct role of time in uniform discrete diffusion: \textit{time tells which subsets of observed tokens are plausibly inherited from the clean sequence}.
A key structural distinction from continuous diffusion lies in the corruption kernel. In continuous diffusion, Gaussian noise continuously perturbs every coordinate, so each corrupted coordinate generally retains a partial contribution from the clean state. In discrete diffusion, by contrast, the state evolves through tokenwise jumps, so only a subset of tokens may change.
Consequently, a corrupted sequence contains tokens directly inherited from the clean sequence alongside tokens generated independently by noise, with the latter carrying no information about their original tokens.
This distinction enables us to clearly understand the specific role of time in uniform discrete diffusion.

Consequently, our analysis reveals a different mechanism through which explicit time conditioning can become unnecessary in uniform discrete diffusion. 
The key distinction is that the role of time itself can be limited for the prediction of $\vz$, rather than time being inferred from $\vz$: for a finite training set, sufficiently separated clean sequences concentrate the empirical posterior onto a small set of compatible hypotheses, making the empirical-optimal predictor nearly invariant to time.
We quantify this through the post-corruption separation margin and directly bound the oracle time sensitivity.

\noindent\textbf{Other related works.}
\citet{liu2025thinkgeneratediscretediffusion} decomposes uniform discrete diffusion into a planner that infers latent corruption identities and a denoiser that predicts clean tokens, highlighting the role of corruption-state uncertainty. Relatedly, \citet{amin2025maskingdiffusionworkscondition} conditions discrete diffusion on the jump schedule and connects uniform diffusion to masking diffusion when corruption-event information is revealed. While these works motivate reasoning about latent corruption states, we instead characterize how such ambiguity induces time dependence in the population-optimal UDM predictor and when this dependence becomes negligible.

\subsection{Various Uniform Discrete Diffusion Frameworks}\label{sec:detailed_background}

We briefly review several representative formulations of uniform discrete
diffusion. We first describe the general CTMC view and its posterior-~\citep{campbell2022continuous} and
score-based parameterizations~\citep{lou2024discrete}. We then introduce the work of \citet{schiff2024simple}, which uses a plug-in UDM parameterization improving empirical performance over these earlier posterior- and score-based parameterizations, and forms the basis of the modern UDM family studied in this work. Finally, we review two recent developments built on this formulation, Diffusion Duality~\citep{sahoo2025diffusionduality} and the leave-one-out parameterization of \citet{gourevitch2026uniformdiffusionmodelsrevisited}.

\noindent\textbf{Posterior and score parameterizations in continuous-time Markov chains.}
Discrete diffusion can be formulated as a continuous-time Markov chain (CTMC)
\citep{campbell2022continuous,lou2024discrete}.
We describe here only the single-token process on
$\calX=\{\ve_1,\ldots,\ve_K\}$, where $\vx,\vz\in\calX$ are categorical
states represented as one-hot vectors. For sequence-level extensions, we refer to the original works
for these constructions. With a slight abuse of notation, write $q_t(\vz\mid\vx)$ for the forward
transition kernel $q_{t|0}(\vz\mid\vx)$ and define its marginal by
\begin{align*}
    p_t(\vz)
    =
    \sum_{\vx\in\calX}
    p_{\mathrm{data}}(\vx)
    q_t(\vz\mid\vx).
\end{align*}
Using the row-vector convention, let
\[
    \vp_t
    =
    \bigl(p_t(\vz)\bigr)_{\vz\in\calX}
    \in \mathbb R^{1\times K},
\]
and let $\mQ_t\in\mathbb R^{K\times K}$ denote the forward transition-rate
matrix. The forward dynamics satisfy
\begin{align*}
    \frac{\mathrm{d}\vp_t}{\mathrm{d}t}
    =
    \vp_t\mQ_t,
\end{align*}
where the off-diagonal entries of $\mQ_t$ are nonnegative and each row sums
to zero. We use this row-vector convention throughout. This is the transpose
of the column-vector convention used by \citet{lou2024discrete}, where
$\mathrm{d}\vp_t/\mathrm{d}t=\mQ_t\vp_t$ and the columns of $\mQ_t$ sum
to zero.

For uniform diffusion, the transition kernel and its corresponding rate
matrix are
\begin{align*}
    \mP_{t|0}
    =
    \alpha_t\mI
    +
    (1-\alpha_t)K^{-1}\vone_K\vone_K^\top,\qquad
    \mQ_t
    =
    -\frac{\alpha_t'}{\alpha_t}
    \left(
        K^{-1}\vone_K\vone_K^\top-\mI
    \right).
\end{align*}
Thus, for two distinct states, the instantaneous transition rate is
$-\alpha_t'/(K\alpha_t)$. Equivalently, one may use the unnormalized
uniform generator
$\vone_K\vone_K^\top-K\mI$, as in \citet{lou2024discrete}, with its
scale absorbed into the scalar noise schedule. Since the uniform generator
is symmetric, the row- and column-vector conventions yield the same numerical
generator matrix in this particular case.

The time reversal is again a CTMC. Under our row-vector convention, its exact
reverse transition rate from $\vz$ to $\vz'\neq\vz$ is
\begin{align*}
    \bar{\mQ}_t(\vz,\vz')
    =
    \mQ_t(\vz',\vz)
    \frac{p_t(\vz')}{p_t(\vz)}.
\end{align*}
The diagonal entries are determined by requiring each row of
$\bar{\mQ}_t$ to sum to zero. Hence, constructing the reverse process amounts
to estimating the marginal ratio $p_t(\vz')/p_t(\vz)$.

The CTMC denoising formulation of \citet{campbell2022continuous} obtains this
ratio indirectly through the clean posterior. By Bayes' rule,
\begin{align*}
    \frac{p_t(\vz')}{p_t(\vz)}
    =
    \sum_{\vx\in\calX}
    \frac{
        q_t(\vz'\mid\vx)
    }{
        q_t(\vz\mid\vx)
    }
    q_{0|t}(\vx\mid\vz).
\end{align*}
Since the clean posterior $q_{0|t}(\vx\mid\vz)$ is generally intractable,
it is replaced by a denoising model
$\vmu_\theta^{\mathrm{Posterior}}(\vx\mid\vz,t)$, yielding
\begin{align*}
    \bar{\mQ}_t^\theta(\vz,\vz')
    =
    \mQ_t(\vz',\vz)
    \sum_{\vx\in\calX}
    \frac{
        q_t(\vz'\mid\vx)
    }{
        q_t(\vz\mid\vx)
    }
    \vmu_\theta^{\mathrm{Posterior}}(\vx\mid\vz,t),
    \qquad
    \vz'\neq\vz.
\end{align*}
Thus, this formulation parameterizes the reverse dynamics through a model of
the clean posterior and trains the resulting reverse process using the
continuous-time NELBO derived from the forward and reverse CTMCs.

SEDD~\citep{lou2024discrete}, in contrast, directly parameterizes the
\emph{concrete score}, i.e., the collection of marginal ratios
\begin{align*}
    s_t(\vz)_{\vz'}
    \coloneqq
    \frac{p_t(\vz')}{p_t(\vz)},
    \qquad
    \vz'\neq\vz.
\end{align*}
Replacing the intractable score by a model
$\vmu_\theta^{\mathrm{Score}}$ gives, in our row-vector convention,
\begin{align*}
    \bar{\mQ}_t^\theta(\vz,\vz')
    =
    \mQ_t(\vz',\vz)
    \vmu_\theta^{\mathrm{Score}}(\vz,t)_{\vz'},
    \qquad
    \vz'\neq\vz.
\end{align*}
SEDD trains these marginal ratios directly using diffusion-weighted denoising
score entropy. Thus, both approaches target the same reverse CTMC but differ
in how it is parameterized: the CTMC denoising formulation models the clean
posterior and obtains the marginal ratio indirectly, whereas SEDD models the
marginal ratio itself as a discrete score.

\noindent\textbf{Plug-in parametrized uniform diffusion models.}
Recent uniform diffusion models largely follow the plug-in parameterization adopted by \citet{schiff2024simple}, which yields empirical improvements over the score- and posterior-based parameterizations of the CTMC formulations described above.
A UDM independently corrupts each token toward the uniform prior:
\begin{align*}
q_t(\vz_t^{i}\mid\vx^{i})
=
\mathrm{Cat}\left(
\vz_t^{i};
\alpha_t\vx^{i}+(1-\alpha_t)K^{-1}\vone_K
\right).
\end{align*}
For $s<t$, the single token posterior $q_{s|0,t}$ follows from Bayes' rule using the $s\to t$ transition and the $s$-marginal:
\begin{align}
\textstyle&q_{s|0,t}(\vz_s^{i}\mid\vx^{i},\vz_t^{i}) =\mathrm{Cat}\left(
\vz_s^{i};
\frac{
\begin{aligned}
&\rho_t\vz_t^{i}\odot\vx^{i}
+\tfrac{\rho_t}{\rho_s}\vz_t^{i}+(1-\tfrac{\rho_t}{\rho_s})(\tfrac{\rho_s\vx^{i}
+\vone_K}{K+\rho_s})
\end{aligned}
}{
\rho_t\langle\vz_t^{i},\vx^{i}\rangle+1
}\label{eq:appendix_posterior}
\right).
\end{align}
To mimic the true posterior, \citet{schiff2024simple} design reverse process with plug-in parametrization:
\begin{align*}
p_{s|t}^{\theta}\!\left(\vz_s^{i}\mid \vz_t\right)
&=q_{s|0,t}(\vz_s^{i}\mid\vx^{i}=\vx_{\theta}^{i}(\vz_t,t),\vz_t)\\
&=
\mathrm{Cat}\!\left(
\vz_s^{i};
\frac{
\rho_t \vz_t^{i} \odot \vx_{\theta}^{i}(\vz_t,t)
+\frac{\rho_t}{\rho_s}\vz_t^{i}+
\left(1-\frac{\rho_t}{\rho_s}\right)
\frac{\rho_s\vx_{\theta}^{i}(\vz_t,t)+\vone_K}{K+\rho_s}
}{
\rho_t
\left\langle
\vz_t^{i},
\vx_{\theta}^{i}(\vz_t,t)
\right\rangle
+1
}
\right).
\end{align*}

The training of UDM uses the variational upper bound (NELBO), and the discrete-time NELBO is:
\begin{align*}\textstyle
    &-\log p_{\theta}(\vx)\\
    &\le\E\bigg[-\log p_{\theta}(\vx\mid\vz_{t(0)})+\sum_i {\mathcal{D}_{\mathrm{KL}}}[q_{s(i)|0,t(i)}(\cdot)\|p_\theta(\vz_{s(i)}\mid\vz_{t(i)})]
    + {\mathcal{D}_{\mathrm{KL}}}[q_{t(T)}(\cdot)\|p_\theta(\vz_{t(T)})]\bigg]
\end{align*}
The prior and reconstruction loss vanish for the continuous-time, and only the diffusion loss remains:
\begin{align*}
\Ls_{\mathrm{UDM}}(\theta)
=\int_0^1\E_{\vx,\vz_t}\sum_{i=1}^L
\frac{-\alpha_t'}{K\alpha_t}
\left[
\frac{K}{\bar{x}_{\theta,\ell_i}^{i}}-\frac{K}{\bar{x}_{\ell_i}^{i}}
+\sum_{j\neq \ell_i}
\frac{\bar{x}_{j}^{i}}{\bar{x}_{\ell_i}^{i}}
\log\frac{\bar{x}_{\theta,\ell_i}^{i}\bar{x}_{j}^{i}}
{\bar{x}_{\theta,j}^{i}\bar{x}_{\ell_i}^{i}}
\right]\mathrm{d}t .
\end{align*}
where $\bar \vnu^{i}\coloneqq K\alpha_t\vnu^{i}+(1-\alpha_t)\vone_K$ and $\ell_i$ denotes non-zero dimension index of $\vz_t^{i}$.

\noindent\textbf{Diffusion duality.}
Diffusion Duality~\citep{sahoo2025diffusionduality} retains the plug-in
parameterization of \citet{schiff2024simple}, but provides an underlying
Gaussian view of the uniform forward process. For each token, consider the
Gaussian diffusion
\begin{align*}
\tilde q_t(\vw_t^{i}\mid\vx^{i};\tilde{\alpha}_t)
=
\calN\!\left(
    \vw_t^{i};
    \tilde{\alpha}_t\vx^{i},
    (1-\tilde{\alpha}_t^2)\mI
\right).
\end{align*}
Diffusion Duality shows that applying $\argmax$ to this Gaussian latent
recovers the marginal of uniform discrete diffusion:
\begin{align*}
q_t\!\left(
    \vz_t^{i}\mid\vx^{i};
    \calT(\tilde{\alpha}_t)
\right)
=
[\argmax]_{\star}\,
\tilde q_t\!\left(
    \vw_t^{i}\mid\vx^{i};
    \tilde{\alpha}_t
\right),
\qquad
\alpha_t=\calT(\tilde{\alpha}_t),
\end{align*}
where $[\argmax]_{\star}$ denotes the pushforward under the $\argmax$
mapping and $\calT$ is the diffusion transformation relating the Gaussian
and uniform noise schedules. Thus, the uniform diffusion marginal can be
viewed as the discrete projection of an underlying Gaussian diffusion.

This duality is further used to construct a low-variance curriculum for
training the same plug-in UDM. Diffusion Duality first rewrites the discrete NELBO exactly using Gaussian latents and a hard argmax projection. Its curriculum then replaces the hard input with a tempered softmax, introducing a lower-variance but biased relaxation that approaches the original discrete-input objective as the temperature decreases. More specifially, Diffusion Duality relaxes this
input using the identity
\begin{align*}
\argmax(\vw_t^{i})
=
\lim_{\tau\to0^+}
\softmax\!\left(\frac{\vw_t^{i}}{\tau}\right).
\end{align*}
During the curriculum, the denoising model therefore receives
$\softmax(\vw_t/\tau)$ instead of the hard $\argmax(\vw_t)$ input.
A positive temperature preserves more information from the Gaussian latent,
making the denoising problem easier and reducing training variance; as
$\tau$ is annealed toward zero, the model progressively approaches the
original discrete UDM training problem. In particular, in their actual experiments, they use the hard $\argmax$ for last half of training, thereby retaining the original UDM's discrete input and NELBO exactly.
 The final model retains the same
plug-in parameterization and is evaluated as a standard uniform diffusion
model.

\noindent\textbf{Leave-one-out parameterization.}
\citet{gourevitch2026uniformdiffusionmodelsrevisited} characterize the target
implicit in the UDM bridge plug-in parameterization. For each position
$i\in[L]$, the population-optimal network prediction is the leave-one-out
(LOO) posterior
\begin{align*}
    \vx_{\star}^{i}(\vz,t)
    =
    \E\left[
        X^{i}
        \,\middle|\,
        Z^{-i}=\vz^{-i},T=t
    \right],
\end{align*}
rather than the full denoising posterior conditioned on the local observation
$Z^{i}=\vz^{i}$. In particular, they show that the UDM NELBO is minimized
when the plug-in predictor satisfies
\begin{align*}
    \vx_{\theta^\star}^{i}(\vz,t)
    =
    \vx_{\star}^{i}(\vz,t)
\end{align*}
for $P_{T,Z}$-almost every $(t,\vz)$. The local observation $\vz^{i}$ can then be incorporated analytically using the known uniform corruption likelihood. By Bayes' rule,
\begin{align*}
    \E\left[
        X^{i}
        \,\middle|\,
        Z=\vz,T=t
    \right]
    =
    \frac{
        \vx_{\star}^{i}(\vz,t)
        \odot
        \left(
            \alpha_t\vz^{i}
            +\frac{1-\alpha_t}{K}\vone_K
        \right)
    }{
        \left\langle
            \vx_{\star}^{i}(\vz,t),
            \alpha_t\vz^{i}
            +\frac{1-\alpha_t}{K}\vone_K
        \right\rangle
    }.
\end{align*}
Thus, the LOO posterior summarizes the information contained in the remaining
corrupted context $Z^{-i}$, while the observation at the target position is
inserted exactly through the known forward corruption process. Replacing the population-optimal LOO posterior $\vx_\star^i$ by the model
prediction $\vx_\theta^i$ gives the corrected posterior denoiser
$\vmu_\theta^{\mathrm{Posterior},i}$. Equivalently, using
$\rho_t=K\alpha_t/(1-\alpha_t)$, this correction can be written in logit form as
\begin{align}
    \vmu_\theta^{\mathrm{Posterior},i}(\vz,t)
    &=
    \softmax\left(
        \log \vx_\theta^{i}(\vz,t)
        +
        \log\left(
            \alpha_t\vz^{i}
            +\frac{1-\alpha_t}{K}\vone_K
        \right)
    \right)\nonumber
    \\
    &=
    \softmax\left(
        \log \vx_\theta^{i}(\vz,t)
        +
        \log(1+\rho_t)\vz^{i}
    \right).\label{eq:loo_parametrization}
\end{align}
The second equality follows because the uniform-channel likelihood differs
between the observed token and all other tokens only by the multiplicative
factor $1+\rho_t$, while any common additive logit term vanishes under
$\softmax$. This corrected posterior can be trained using the usual denoising
cross-entropy,
\begin{align*}
    \Ls_{\mathrm{LOO+CE}}(\theta)
    =
    -\int_0^1
    \E_{\vx,\vz_t}
    \sum_{i=1}^{L}
    \log
    \left\langle
        \vx^{i},
        \vmu_\theta^{\mathrm{Posterior},i}(\vz_t,t)
    \right\rangle
    \mathrm{d}t .
\end{align*}
At the unrestricted population optimum, the corrected posterior satisfies
$\vmu_{\theta^\star}^{\mathrm{Posterior},i}(\vz,t)
=
\E[X^{i}\mid Z=\vz,T=t]$, and, crucially, the underlying predictor satisfies
\begin{align*}
    \vx_{\theta^\star}^{i}(\vz,t)
    =
    \vx_{\star}^{i}(\vz,t)
    =
    \E\left[
        X^{i}
        \,\middle|\,
        Z^{-i}=\vz^{-i},T=t
    \right].
\end{align*}
Thus, the corrected cross-entropy objective provides a convenient way to
train the same LOO predictor required by the UDM plug-in parameterization.
Once trained, $\vx_\theta$ is plugged into the original UDM bridge
$p_{s|t}^{\theta}$ to define the reverse transitions used for generation.

\section{Omitted Proofs}\label{sec:proof}
In this section, we provide omitted proofs in the main paper. Throughout the proofs, we consider $0<t<1$ with $0<\alpha_t<1$.
Conditional expectations on events of zero data probability are
assigned arbitrary, time-independent probability vectors, since
their corresponding mixture weights are zero.

\subsection{Proof of Proposition~\ref{prop:optimum}}
\optimum*
\begin{proof}
    \citet{gourevitch2026uniformdiffusionmodelsrevisited} showed that the
    unrestricted population minimizer of the bridge plug-in UDM NELBO is the
    leave-one-out posterior. Therefore,
    \begin{align*}
        \vx_{\theta^\star}^{i}(\vz,t)
        &=
        \vx_{\star}^{i}(\vz,t)
        =
        \E
        \left[
            X^{i}
            \,\middle|\,
            Z^{-i}=\vz^{-i},T=t
        \right].
    \end{align*}
    By Bayes' rule and marginalization,
    \begin{align}
        \vx_{\star}^{i}(\vz,t)
        &=
        \frac{
            \E_{p_{\mathrm{data}}}
            \left[
                X^{i}
                q_t\!\left(
                    \vz^{-i}
                    \mid
                    X^{-i}
                \right)
            \right]
        }{
            \E_{p_{\mathrm{data}}}
            \left[
                q_t\!\left(
                    \vz^{-i}
                    \mid
                    X^{-i}
                \right)
            \right]
        }.
        \label{eq:loo-bayes}
    \end{align}

    Since the uniform forward process factorizes across positions,
    \begin{align}
        q_t\!\left(
            \vz^{-i}
            \mid
            X^{-i}
        \right)
        &=
        \prod_{j\in[L]\setminus\{i\}}
        \left(
            \alpha_t
            \ind\!\left\{
                X^{j}=\vz^{j}
            \right\}
            +
            \frac{1-\alpha_t}{K}
        \right)
        \notag\\
        &=
        \left(
            \frac{1-\alpha_t}{K}
        \right)^{L-1}
        \prod_{j\in[L]\setminus\{i\}}
        \left(
            1+
            \rho_t
            \ind\!\left\{
                X^{j}=\vz^{j}
            \right\}
        \right)
        \notag\\
        &=
        \left(
            \frac{1-\alpha_t}{K}
        \right)^{L-1}
        \sum_{S\subseteq[L]\setminus\{i\}}
        \rho_t^{|S|}
        \ind\!\left\{
            X[S]=\vz[S]
        \right\},
        \label{eq:subset-expansion}
    \end{align}
    where the last equality follows by expanding the product over all subsets
    of $[L]\setminus\{i\}$.

    Substituting \eqref{eq:subset-expansion} into
    \eqref{eq:loo-bayes} and canceling the common factor
    $\bigl((1-\alpha_t)/K\bigr)^{L-1}$ gives
    \begin{align*}
        \vx_{\star}^{i}(\vz,t)
        &=
        \frac{
            \displaystyle
            \sum_{S\subseteq[L]\setminus\{i\}}
            \rho_t^{|S|}
            \E_{p_{\mathrm{data}}}
            \left[
                X^{i}
                \ind\!\left\{
                    X[S]=\vz[S]
                \right\}
            \right]
        }{
            \displaystyle
            \sum_{S\subseteq[L]\setminus\{i\}}
            \rho_t^{|S|}
            \Pr
            \left(
                X[S]=\vz[S]
            \right)
        }.
    \end{align*}

    Using
    \begin{align*}
        &\E_{p_{\mathrm{data}}}
        \left[
            X^{i}
            \ind\!\left\{
                X[S]=\vz[S]
            \right\}
        \right]
        =
        \Pr
        \left(
            X[S]=\vz[S]
        \right)
        \E
        \left[
            X^{i}
            \,\middle|\,
            X[S]=\vz[S]
        \right],
    \end{align*}
    we obtain
    \begin{align}
        \vx_{\star}^{i}(\vz,t)
        &=
        \sum_{S\subseteq[L]\setminus\{i\}}
        \frac{
            \rho_t^{|S|}
            \Pr
            \left(
                X[S]=\vz[S]
            \right)
        }{
            \displaystyle
            \sum_{S'\subseteq[L]\setminus\{i\}}
            \rho_t^{|S'|}
            \Pr
            \left(
                X[S']=\vz[S']
            \right)
        }
        \E
        \left[
            X^{i}
            \,\middle|\,
            X[S]=\vz[S]
        \right].
        \label{eq:subset-mixture}
    \end{align}

    It remains to identify the normalized coefficient in
    \eqref{eq:subset-mixture}. Interpret $S$ as the set of context positions
    generated through the copy branch of the uniform forward channel.
    For a fixed $S$, the joint probability of this copy-set hypothesis and
    the observation $Z^{-i}=\vz^{-i}$ is
    \begin{align*}
        \Pr
        \left(
            S,\,
            Z^{-i}=\vz^{-i}
            \,\middle|\,
            T=t
        \right)
        &=
        \alpha_t^{|S|}
        \left(
            \frac{1-\alpha_t}{K}
        \right)^{L-1-|S|}
        \Pr
        \left(
            X[S]=\vz[S]
        \right)
        \notag\\
        &=
        \left(
            \frac{1-\alpha_t}{K}
        \right)^{L-1}
        \rho_t^{|S|}
        \Pr
        \left(
            X[S]=\vz[S]
        \right).
    \end{align*}
    Therefore, normalizing over all
    $S\subseteq[L]\setminus\{i\}$ gives
    \begin{align}
        \Pr
        \left(
            S
            \,\middle|\,
            Z^{-i}=\vz^{-i},T=t
        \right)
        =
        \frac{
            \rho_t^{|S|}
            \Pr
            \left(
                X[S]=\vz[S]
            \right)
        }{
            \displaystyle
            \sum_{S'\subseteq[L]\setminus\{i\}}
            \rho_t^{|S'|}
            \Pr
            \left(
                X[S']=\vz[S']
            \right)
        }.
        \label{eq:copy-set-posterior}
    \end{align}
    Hence,
    \begin{align*}
        \Pr
        \left(
            S
            \,\middle|\,
            Z^{-i}=\vz^{-i},T=t
        \right)
        \propto
        \rho_t^{|S|}
        \Pr
        \left(
            X[S]=\vz[S]
        \right),
    \end{align*}
    and substituting \eqref{eq:copy-set-posterior} into
    \eqref{eq:subset-mixture} yields
    \begin{align*}
        \vx_{\theta^\star}^{i}(\vz,t)
        &=
        \sum_{S\subseteq[L]\setminus\{i\}}
        \Pr
        \left(
            S
            \,\middle|\,
            Z^{-i}=\vz^{-i},T=t
        \right)
        \E
        \left[
            X^{i}
            \,\middle|\,
            X[S]=\vz[S]
        \right],
    \end{align*}
    which proves the claim.
\end{proof}
\subsection{Proof of Theorem~\ref{thm:exponential_reliability}}
\maskcount*
\begin{proof}
By the law of total expectation,
\begin{align}
    \vx_{\star}^{i}(\vz,t)
    =
    \sum_{m=0}^{L-1}
    \Pr
    \left(
        M^i=m
        \,\middle|\,
        Z^{-i}=\vz^{-i},T=t
    \right)
    \times
    \E
    \left[
        X^{i}
        \,\middle|\,
        Z^{-i}=\vz^{-i},M^i=m,T=t
    \right].
    \label{eq:match_count_decomposition}
\end{align}
Therefore, it remains to show that
\begin{align*}
    \Pr
    \left(
        M^i=m
        \,\middle|\,
        Z^{-i}=\vz^{-i},T=t
    \right)
    &\propto
    (1+\rho_t)^m\Pr(M^i(X,\vz)=m),
\end{align*}
and
\begin{align*}
    &\E
    \left[
        X^{i}
        \,\middle|\,
        Z^{-i}=\vz^{-i},M^i=m,T=t
    \right]
    =
    \E
    \left[
        X^{i}
        \,\middle|\,
        Z^{-i}=\vz^{-i},M^i=m
    \right].
\end{align*}

\noindent\textbf{Posterior weighting.}
For any clean sequence $\vx'$, the uniform forward channel gives
\begin{align}
    q_t
    \left(
        \vz^{-i}
        \,\middle|\,
        \vx'^{-i}
    \right)
    =
    \prod_{j\in[L]\setminus\{i\}}
    \left[
        \alpha_t
        \ind
        \left\{
            \vx'^{j}=\vz^{j}
        \right\}
        +
        \frac{1-\alpha_t}{K}
    \right]
    =
    \left(
        \frac{1-\alpha_t}{K}
    \right)^{L-1}
    (1+\rho_t)^{M^i(\vx',\vz)}.
    \label{eq:match_count_likelihood}
\end{align}
Hence,
\begin{align*}
    \Pr
    \left(
        M^i=m,
        Z^{-i}=\vz^{-i}
        \,\middle|\,
        T=t
    \right)
    &=
    \sum_{\substack{
        \vx'\in\calX^L\\
        M^i(\vx',\vz)=m
    }}
    p_{\mathrm{data}}(\vx')
    q_t
    \left(
        \vz^{-i}
        \,\middle|\,
        \vx'^{-i}
    \right)
    \notag\\
    &=
    \left(
        \frac{1-\alpha_t}{K}
    \right)^{L-1}
    (1+\rho_t)^m
    \sum_{\substack{
        \vx'\in\calX^L\\
        M^i(\vx',\vz)=m
    }}
    p_{\mathrm{data}}(\vx')
    \notag\\
    &=
    \left(
        \frac{1-\alpha_t}{K}
    \right)^{L-1}
    \Pr(M^i(X,\vz)=m)
    (1+\rho_t)^m.
\end{align*}
Therefore,
\begin{align*}
    \Pr
    \left(
        M^i=m
        \,\middle|\,
        Z^{-i}=\vz^{-i},T=t
    \right)
    &=
    \frac{
        \Pr
        \left(
            M^i=m,
            Z^{-i}=\vz^{-i}
            \,\middle|\,
            T=t
        \right)
    }{
        \sum_{r=0}^{L-1}\Pr
        \left(
            M^i=r,
            Z^{-i}=\vz^{-i}
            \,\middle|\,
            T=t
        \right)
    }
    \notag\\
    &=
    \frac{
        \Pr(M^i(X,\vz)=m)(1+\rho_t)^m
    }{
        \displaystyle
        \sum_{r=0}^{L-1}
        \Pr(M^i(X,\vz)=r)(1+\rho_t)^r
    }.
\end{align*}
and hence
\[
\Pr
    \left(
        M^i=m
        \,\middle|\,
        Z^{-i}=\vz^{-i},T=t
    \right)\propto \Pr(M^i(X,\vz)=m)(1+\rho_t)^m.
\]

\noindent\textbf{Time-independence of the conditional prediction.}
For any $\vx'$ satisfying $M^i(\vx',\vz)=m$, Bayes' rule and
\eqref{eq:match_count_likelihood} give
\begin{align*}
    \Pr
    \left(
        X=\vx'
        \,\middle|\,
        Z^{-i}=\vz^{-i},M^i=m,T=t
    \right)
    &=
    \frac{
        p_{\mathrm{data}}(\vx')
        q_t
        \left(
            \vz^{-i}
            \,\middle|\,
            \vx'^{-i}
        \right)
    }{
        \displaystyle
        \sum_{\substack{
            \vx''\in\calX^L\\
            M^i(\vx'',\vz)=m
        }}
        p_{\mathrm{data}}(\vx'')
        q_t
        \left(
            \vz^{-i}
            \,\middle|\,
            \vx''^{-i}
        \right)
    }
    \notag\\
    &=
    \frac{
        p_{\mathrm{data}}(\vx')
    }{
        \displaystyle
        \sum_{\substack{
            \vx''\in\calX^L\\
            M^i(\vx'',\vz)=m
        }}
        p_{\mathrm{data}}(\vx'')
    }
    =
    \frac{
        p_{\mathrm{data}}(\vx')
    }{
        \Pr(M^i(X,\vz)=m)
    },
\end{align*}
where the second equality follows because all sequences satisfying $M^i(\vx',\vz)=m$ have the same uniform-channel likelihood, so the common factor cancels under normalization.

The right-hand side is independent of $t$. Therefore,
\begin{align*}
    X^{i}
    \perp
    T
    \mid
    Z^{-i},M^i,
\end{align*}
and consequently
\begin{align*}
    \E
    \left[
        X^{i}
        \,\middle|\,
        Z^{-i}=\vz^{-i},M^i=m,T=t
    \right]
    =
    \E
    \left[
        X^{i}
        \,\middle|\,
        Z^{-i}=\vz^{-i},M^i=m
    \right]
    =
    \mathbf h_m^{i}(\vz).
\end{align*}

Substituting the two results into
\eqref{eq:match_count_decomposition} gives
\begin{align*}
    \vx_{\star}^{i}(\vz,t)
    =
    \sum_{m=0}^{L-1}
    w_m^i(\vz,t)\mathbf h_m^{i}(\vz),
\end{align*}
with
\begin{align*}
    w_m^i(\vz,t)
    \propto
    \Pr(M^i(X,\vz)=m)(1+\rho_t)^m,
\end{align*}
which proves the claim.
\end{proof}

\subsection{Proof of Proposition~\ref{prop:exact_temporal_sensitivity}}
\derivative*

\begin{proof}
By Theorem~\ref{thm:exponential_reliability},
\begin{align*}
    \vx_{\star}^{i}(\vz,t)
    =
    \sum_{m=0}^{L-1}
    w_m^i(\vz,t)\mathbf h_m^{i}(\vz),
\end{align*}
where, writing $\eta_t=\log(1+\rho_t)$,
\begin{align}
    w_m^i(\vz,t)
    =
    \frac{
        \Pr(M^i(X,\vz)=m)e^{m\eta_t}
    }{
        \displaystyle
        \sum_{r=0}^{L-1}
        \Pr(M^i(X,\vz)=r)e^{r\eta_t}
    }.
    \label{eq:wm_exponential_form}
\end{align}
Since $\mathbf h_m^{i}(\vz)$ is independent of $t$, all dependence on
$\eta_t$ is contained in $w_m^i(\vz,t)$. Differentiating
\eqref{eq:wm_exponential_form} gives
\begin{align*}
    \frac{\partial}{\partial\eta_t}
    w_m^i(\vz,t)
    &=
    w_m^i(\vz,t)
    \left(
        m-
        \sum_{r=0}^{L-1}r\,w_r^i(\vz,t)
    \right)
    \notag\\
    &=
    w_m^i(\vz,t)
    \left(
        m-\E_{t,\vz}[M^i]
    \right).
\end{align*}
Therefore,
\begin{align*}
    \frac{\partial}{\partial\eta_t}
    \vx_{\star}^{i}(\vz,t)
    &=
    \sum_{m=0}^{L-1}
    \mathbf h_m^{i}(\vz)
    w_m^i(\vz,t)
    \left(
        m-\E_{t,\vz}[M^i]
    \right)
    \notag\\
    &=
    \E_{t,\vz}
    \left[
        \mathbf h_{M^i}^{i}(\vz)M^i
    \right]
    -
    \E_{t,\vz}
    \left[
        \mathbf h_{M^i}^{i}(\vz)
    \right]
    \E_{t,\vz}[M^i]
    \notag\\
    &=
    \operatorname{Cov}_{t,\vz}
    \left(
        \mathbf h_{M^i}^{i}(\vz),
        M^i
    \right),
\end{align*}
which proves the claim.

Now fix a candidate token $\ve_k$ such that
$\vx_{\star,k}^{i}(\vz,t)>0$. Taking the $k$-th component of the
previous expression,
\begin{align}
    \frac{\partial}{\partial\eta_t}
    \log
    \vx_{\star,k}^{i}(\vz,t)
    &=
    \frac{
        \displaystyle
        \sum_{m=0}^{L-1}
        w_m^i(\vz,t)
        h_{m,k}^{i}(\vz)m
    }{
        \displaystyle
        \sum_{m=0}^{L-1}
        w_m^i(\vz,t)
        h_{m,k}^{i}(\vz)
    }
    -
    \E_{t,\vz}[M^i],
    \label{eq:log_derivative_before_bayes}
\end{align}
where $h_{m,k}^{i}(\vz)$ denotes the $k$-th component of
$\mathbf h_m^{i}(\vz)$.

By Theorem~\ref{thm:exponential_reliability},
$X^{i}\perp T\mid Z^{-i},M^i$, and hence we can include $t$ into the distribution:
\begin{align*}
    h_{m,k}^{i}(\vz)
    &=
    \Pr
    \left(
        X^{i}=\ve_k
        \,\middle|\,
        M^i=m,
        Z^{-i}=\vz^{-i},
        T=t
    \right).
\end{align*}
Together with
\begin{align*}
    w_m^i(\vz,t)
    =
    \Pr
    \left(
        M^i=m
        \,\middle|\,
        Z^{-i}=\vz^{-i},
        T=t
    \right),
\end{align*}
Bayes' rule gives
\begin{align*}
    \frac{
        w_m^i(\vz,t)h_{m,k}^{i}(\vz)
    }{
        \vx_{\star,k}^{i}(\vz,t)
    }
    =
    \Pr
    \left(
        M^i=m
        \,\middle|\,
        X^{i}=\ve_k,
        Z^{-i}=\vz^{-i},
        T=t
    \right).
\end{align*}
Consequently, the first term in
\eqref{eq:log_derivative_before_bayes} is precisely
\begin{align*}
    \E_{t,\vz}
    \left[
        M^i
        \,\middle|\,
        X^{i}=\ve_k
    \right],
\end{align*}
and therefore
\begin{align*}
    \frac{\partial}{\partial\eta_t}
    \log
    \vx_{\star,k}^{i}(\vz,t)
    =
    \E_{t,\vz}
    \left[
        M^i
        \,\middle|\,
        X^{i}=\ve_k
    \right]
    -
    \E_{t,\vz}[M^i].
\end{align*}
This proves \eqref{eq:single_candidate_reliability_sensitivity}.
\end{proof}

\subsection{Proof of Proposition~\ref{lem:delta-seperable}}\label{sec:formal_delta}
We here provide a formal version of Proposition~\ref{lem:delta-seperable} in the main paper and the proof under the empirical data distribution setting.
\begin{asm}[Finite-data setting]\label{asm:finite}
The data distribution is the uniform empirical distribution\footnote{Note that every theoretical result in our paper holds within Assumption~\ref{asm:finite}; see Appendix~\ref{sec:assumption} for why.}
$\real=\frac{1}{N}\sum_{n=1}^N\delta_{\vx_n}$ over $N$ sequences
$\vx_1,\ldots,\vx_N\in\calX^L$ where $|\calX|=K$.
\end{asm}
\setcounter{formalProposition}{2}
\begin{formalProposition}[$\Delta$-separable event under UDM training]
\label{lem:delta-separable-formal}
Under Assumption~\ref{asm:finite}, fix
$\vx\in\operatorname{supp}(\real)$ and $i\in[L]$, and suppose
$d(\vx)\ge2$.
Let $T\sim\operatorname{Unif}([0,1])$ and, conditioned on $T=t$,
let $Z\sim q_t(\cdot\mid\vx)$ under the linear schedule
$\alpha_t=1-t$.
Then, for any integer $\Delta\ge1$,
\begin{equation}
\Pr\!\left(
Z
\text{ is $\Delta$-separable w.r.t.\ $\vx$ at $i$}
\right)
\ge
\max\left\{
0,\,
1-
\frac{
\log(N-1)+\Delta
+
(e+e^{-1}-2)\frac{L-1}{K}
}{
(1-e^{-1})(d(\vx)-1)
}
\right\}.\label{eq:exact_delta_prob_bound}
\end{equation}
\end{formalProposition}
Thus, as given in the informal statement of the main paper, if $K\gg L$ and $d(\vx)$ is sufficiently large, the approximate bound
\[
\Pr\!\left(
Z
\text{ is $\Delta$-separable w.r.t. $\vx$ at $i$}
\right)
\gtrsim
1-
\frac{\log N+\Delta}
{(1-e^{-1})d(\vx)}
\]
holds. Note that we used \eqref{eq:exact_delta_prob_bound} to measure quantitative bound in Section~\ref{sec:numerical_analysis}. We now provide the proof of formal statement:

\begin{proof}
Fix any
$\vx'\in\operatorname{supp}(\real)$ satisfying
$\vx'^{i}\neq\vx^{i}$, and let
\[
\mathcal H_i(\vx,\vx')
\coloneqq
\left\{
j\in[L]\setminus\{i\}:
\vx'^{j}\neq\vx^{j}
\right\}.
\]
Since $\vx'^{i}\neq\vx^{i}$ and the Hamming distance between
$\vx$ and $\vx'$ is at least $d(\vx)$,
\begin{align}
\left|\mathcal H_i(\vx,\vx')\right|
&=
\sum_{j\in[L]\setminus\{i\}}
\ind
\left\{
\vx'^{j}\neq\vx^{j}
\right\}
\ge
d(\vx)-1.
\label{eq:hamming_context_lower}
\end{align}

Condition on $T=t$. For each
$j\in\mathcal H_i(\vx,\vx')$, define
\[
Y_j
\coloneqq
\ind\left\{Z^{j}=\vx^{j}\right\}
-
\ind\left\{Z^{j}=\vx'^{j}\right\}.
\]
For $j\notin\mathcal H_i(\vx,\vx')$, $\vx$ and $\vx'$ contribute equally
to the match count. Therefore,
\begin{align}
M^i(\vx,Z)-M^i(\vx',Z)
=
\sum_{j\in\mathcal H_i(\vx,\vx')}Y_j.
\label{eq:match_gap_sum}
\end{align}

Under the linear schedule $\alpha_t=1-t$, the uniform forward kernel gives,
for every $j\in\mathcal H_i(\vx,\vx')$,
\begin{align*}
\Pr(Y_j=1\mid T=t)
&=
1-t+\frac{t}{K},
\\
\Pr(Y_j=0\mid T=t)
&=
\frac{t(K-2)}{K},
\\
\Pr(Y_j=-1\mid T=t)
&=
\frac{t}{K}.
\end{align*}
Moreover, the variables $\{Y_j\}_{j\in\mathcal H_i(\vx,\vx')}$ are
independent conditioned on $T=t$. Hence,
\begin{align}
\E\left[e^{-Y_j}\mid T=t\right]
&=
\left(
1-t+\frac{t}{K}
\right)e^{-1}
+
\frac{t(K-2)}{K}
+
\frac{t}{K}e
\notag\\
&=
1
-
(1-e^{-1})(1-t)
+
(e+e^{-1}-2)\frac{t}{K}\nonumber\\
&\le
\exp\left(
-(1-e^{-1})(1-t)
+
(e+e^{-1}-2)\frac{t}{K}
\right),
\label{eq:single_match_mgf}
\end{align}
where the last inequality comes from $1+u\le e^{u}$.

The sequence $Z$ fails to be $\Delta$-separable from this particular
$\vx'$ whenever
\[
M^i(\vx',Z)>M^i(\vx,Z)-\Delta.
\]
Since the match counts are integer-valued, this is equivalent to
\[
M^i(\vx,Z)-M^i(\vx',Z)\le\Delta-1.
\]
Therefore, by Markov's inequality and
\eqref{eq:match_gap_sum},
\begin{align*}
&\Pr\left(
M^i(\vx,Z)-M^i(\vx',Z)\le\Delta-1
\,\middle|\,
T=t
\right)
\notag\\
&\qquad=
\Pr\left(
\exp\!\left(
-\bigl(M^i(\vx,Z)-M^i(\vx',Z)\bigr)
\right)
\ge e^{-(\Delta-1)}
\,\middle|\,
T=t
\right)
\notag\\
&\qquad\le
e^{\Delta-1}
\E\left[
\exp\!\left(
-\bigl(M^i(\vx,Z)-M^i(\vx',Z)\bigr)
\right)
\,\middle|\,
T=t
\right]
\notag\\
&\qquad=
e^{\Delta-1}
\E\left[
\exp\left(
-\sum_{j\in\mathcal H_i(\vx,\vx')}Y_j
\right)
\,\middle|\,
T=t
\right]
\notag\\
&\qquad=
e^{\Delta-1}
\prod_{j\in\mathcal H_i(\vx,\vx')}
\E\left[e^{-Y_j}\mid T=t\right],
\end{align*}
Plugging \eqref{eq:single_match_mgf} into the above inequality gives:
\begin{align*}
&\Pr\left(
M^i(\vx,Z)-M^i(\vx',Z)\le\Delta-1
\,\middle|\,
T=t
\right)
\notag\\
&\qquad\le
e^{\Delta-1}
\prod_{j\in\mathcal H_i(\vx,\vx')}
\exp\left(
-(1-e^{-1})(1-t)
+
(e+e^{-1}-2)\frac{t}{K}
\right)
\notag\\
&\qquad=
\exp\Bigg(
\Delta-1
-
(1-e^{-1})(1-t)
\left|\mathcal H_i(\vx,\vx')\right|
+
(e+e^{-1}-2)
\frac{t\left|\mathcal H_i(\vx,\vx')\right|}{K}
\Bigg)\\
&\qquad\le
\exp\Bigg(
\Delta-1
-
(1-e^{-1})(1-t)(d(\vx)-1)
+
(e+e^{-1}-2)\frac{L-1}{K}
\Bigg).
\end{align*}
where the last inequality uses
\eqref{eq:hamming_context_lower},
$t\le1$, and
$\left|\mathcal H_i(\vx,\vx')\right|\le L-1$.

There are at most $N-1$ empirical sequences distinct from $\vx$.
Thus, by a union bound over all
$\vx'\in\operatorname{supp}(\real)$ satisfying
$\vx'^{i}\neq\vx^{i}$,
\begin{align}
&\Pr\left(
Z\text{ is not $\Delta$-separable w.r.t.\ $\vx$ at $i$}
\,\middle|\,
T=t
\right)
\notag\\
&\qquad\le
\min\Bigg\{
1,\,
(N-1)
\exp\Bigg(
\Delta-1
-
(1-e^{-1})(1-t)(d(\vx)-1)
+
(e+e^{-1}-2)\frac{L-1}{K}
\Bigg)
\Bigg\}.
\label{eq:conditional_separation_failure}
\end{align}

It remains to average over the training time
$T\sim\operatorname{Unif}([0,1])$.
For any $A\ge1$ and $B>0$,
\begin{align}
\int_0^1
\min\left\{
1,\,
Ae^{-B(1-t)}
\right\}
\mathrm{d}t
\le
\min\left\{
1,\,
\frac{\log A+1}{B}
\right\}.
\label{eq:truncated_exponential_integral}
\end{align}
Indeed, if $\log A\ge B$, the bound follows immediately since the
integral is at most one. Otherwise, splitting the integral at
$t=1-\log A/B$ gives
\begin{align}
\int_0^1
\min\left\{
1,\,
Ae^{-B(1-t)}
\right\}
\mathrm{d}t
=
\frac{\log A}{B}
+
\frac{1-Ae^{-B}}{B}
\notag \le
\frac{\log A+1}{B}.
\end{align}

Applying \eqref{eq:truncated_exponential_integral} to
\eqref{eq:conditional_separation_failure} with
\[
A
=
(N-1)
\exp\left(
\Delta-1
+
(e+e^{-1}-2)\frac{L-1}{K}
\right)
\]
and
\[
B
=
(1-e^{-1})(d(\vx)-1),
\]
we obtain
\begin{align*}
&\Pr\left(
Z\text{ is not $\Delta$-separable w.r.t.\ $\vx$ at $i$}
\right)
\le
\min\left\{
1,\,
\frac{
\log(N-1)+\Delta
+
(e+e^{-1}-2)\frac{L-1}{K}
}{
(1-e^{-1})(d(\vx)-1)
}
\right\}.
\end{align*}
Taking the complement yields
\begin{equation*}
\Pr\left(
Z
\text{ is $\Delta$-separable w.r.t.\ $\vx$ at $i$}
\right)
\ge
\max\left\{
0,\,
1-
\frac{
\log(N-1)+\Delta
+
(e+e^{-1}-2)\frac{L-1}{K}
}{
(1-e^{-1})(d(\vx)-1)
}
\right\},
\end{equation*}
which proves the claim. 
\end{proof}

\subsection{Proof of Theorem~\ref{thm:finitefisher}}
We here provide a formal version of Theorem~\ref{thm:finitefisher}
in the main paper under the empirical data distribution setting.
We first define the Fisher sensitivity with respect to the noise-schedule
parameter $\alpha_t$:
\[
\mathcal I_{\alpha_t}^{i}(\vz)
\coloneqq
\sum_{k:\,\vx_{\star,k}^{i}(\vz,t)>0}
\vx_{\star,k}^{i}(\vz,t)
\left(
\frac{\partial}{\partial \alpha_t}
\log \vx_{\star,k}^{i}(\vz,t)
\right)^2.
\]

\setcounter{formalthm}{1}
\begin{formalthm}[Fisher sensitivity with finite data]
Suppose that $\alpha_t$ is strictly decreasing in $t$.
Under the finite data assumption~\ref{asm:finite}, for $\Delta$-separable $\vz$ with respect to $\vx$ at coordinate $i$,
\[
\mathcal{I}_{\alpha_t}^i(\vz)
\le
(N-1)(L-1)^2 K^2
\frac{(1-\alpha_t)^{\Delta-2}}
{\left(1+(K-1)\alpha_t\right)^{\Delta+2}}.
\]
Moreover, for $\Delta\ge2$, the right-hand side is strictly increasing
in $t\in(0,1)$.
\end{formalthm}

The bound admits the same interpretation as its time-based counterpart.
What matters for denoising is how much the oracle prediction changes
per unit change in the corruption level, which is measured directly by
$\mathcal I_{\alpha_t}^i(\vz)$.
The theorem shows that as the corruption level increases
(i.e., as $\alpha_t$ decreases), the upper bound on this sensitivity
becomes progressively weaker.
Thus, time dependence is most weakly controlled in the high-noise regime. 

For ease of interpretation, the main text specializes this result to the
linear schedule $\alpha_t=1-t$ and expresses the Fisher sensitivity with
respect to $t$. Under this schedule, the theorem reduces to
\[
\mathcal I_t^{i}(\vz)
\le
(N-1)(L-1)^2
\frac{
K^2 t^{\Delta-2}
}{
\bigl(K-(K-1)t\bigr)^{\Delta+2}
}.
\]
Using $(N-1)(L-1)^2 < NL^2$ gives the simpler bound stated in the
main paper:
\[
\mathcal I_t^{i}(\vz)
\le
NL^2
\frac{
K^2 t^{\Delta-2}
}{
\bigl(K-(K-1)t\bigr)^{\Delta+2}
}.
\]
We now prove the formal version of Theorem~\ref{thm:finitefisher}.

\begin{proof}
By Proposition~\ref{prop:exact_temporal_sensitivity}, with
$\eta_t=\log(1+\rho_t)$, for every $k$ such that
$\vx_{\star,k}^{i}(\vz,t)>0$,
\begin{align*}
    \frac{\partial}{\partial \alpha_t}
    \log \vx_{\star,k}^{i}(\vz,t)
    &=
    \frac{\mathrm{d}\eta_t}{\mathrm{d}\alpha_t}
    \left(
        \E_{t,\vz}
        \left[
            M^i
            \,\middle|\,
            X^{i}=\ve_k
        \right]
        -
        \E_{t,\vz}[M^i]
    \right).
\end{align*}
Therefore, by the definition of
$\mathcal I_{\alpha_t}^{i}(\vz)$,
\begin{align}
    \mathcal I_{\alpha_t}^{i}(\vz)
    &=
    \left(
        \frac{\mathrm{d}\eta_t}{\mathrm{d}\alpha_t}
    \right)^2
    \sum_{k:\,\vx_{\star,k}^{i}(\vz,t)>0}
    \vx_{\star,k}^{i}(\vz,t)
    \left(
        \E_{t,\vz}
        \left[
            M^i
            \,\middle|\,
            X^{i}=\ve_k
        \right]
        -
        \E_{t,\vz}[M^i]
    \right)^2
    \nonumber\\
    &=
    \left(
        \frac{\mathrm{d}\eta_t}{\mathrm{d}\alpha_t}
    \right)^2
    \E_{t,\vz}
    \left[
        \left(
            \E_{t,\vz}
            \left[
                M^i
                \,\middle|\,
                X^{i}
            \right]
            -
            \E_{t,\vz}[M^i]
        \right)^2
    \right]
    \nonumber\\
    &=
    \left(
        \frac{\mathrm{d}\eta_t}{\mathrm{d}\alpha_t}
    \right)^2
    \operatorname{Var}_{t,\vz}
    \left(
        \E_{t,\vz}
        \left[
            M^i
            \,\middle|\,
            X^{i}
        \right]
    \right).
    \label{eq:finite_fisher_variance}
\end{align}

For any choice of center, the variance is upper bounded by the
corresponding mean squared deviation. We therefore choose the conditional
mean associated with the source token $\vx^i$,
$\E_{t,\vz}[M^i\mid X^i=\vx^i]$, where $\vx$ is the clean sequence from
which $\vz$ was sampled. This choice is convenient because the deviation
vanishes whenever $X^i=\vx^i$. Moreover, since $0\le M^i\le L-1$, all
conditional expectations below lie in $[0,L-1]$. Hence,
\begin{align}
    \operatorname{Var}_{t,\vz}
    \left(
        \E_{t,\vz}
        \left[
            M^i
            \,\middle|\,
            X^{i}
        \right]
    \right)
    &\le
    \E_{t,\vz}
    \left[
        \left(
            \E_{t,\vz}
            \left[
                M^i
                \,\middle|\,
                X^{i}
            \right]
            -
            \E_{t,\vz}
            \left[
                M^i
                \,\middle|\,
                X^{i}=\vx^{i}
            \right]
        \right)^2
    \right]
    \notag\\
    &\le
    (L-1)^2
    \Pr
    \left(
        X^{i}\neq \vx^{i}
        \,\middle|\,
        Z^{-i}=\vz^{-i},T=t
    \right).
    \label{eq:finite_fisher_wrong_target}
\end{align}

It remains to bound the posterior probability of a different target token.
Under Assumption~\ref{asm:finite},
\[
\real
=
\frac{1}{N}\sum_{n=1}^{N}\delta_{\vx_n}.
\]
From \eqref{eq:match_count_likelihood},
\begin{align*}
    q_t
    \left(
        \vz^{-i}
        \,\middle|\,
        \vx_n^{-i}
    \right)
    =
    \left(
        \frac{1-\alpha_t}{K}
    \right)^{L-1}
    (1+\rho_t)^{M^i(\vx_n,\vz)}.
\end{align*}
Hence, since the empirical prior is uniform,
\begin{align}
    \Pr
    \left(
        X^{i}\neq \vx^{i}
        \,\middle|\,
        Z^{-i}=\vz^{-i},T=t
    \right)
    =
    \frac{
        \displaystyle
        \sum_{\substack{
            n:\,\vx_n^{i}\neq\vx^{i}
        }}
        (1+\rho_t)^{M^i(\vx_n,\vz)}
    }{
        \displaystyle
        \sum_{r=1}^{N}
        (1+\rho_t)^{M^i(\vx_r,\vz)}
    }.
    \label{eq:finite_wrong_target_posterior}
\end{align}
By $\Delta$-separability, every sequence satisfying
$\vx_n^{i}\neq\vx^{i}$ obeys
\begin{align*}
    M^i(\vx_n,\vz)
    \le
    M^i(\vx,\vz)-\Delta.
\end{align*}
Moreover, the denominator of
\eqref{eq:finite_wrong_target_posterior} contains the contribution of the
sampled sequence $\vx$. Thus,
\begin{align}
    \Pr
    \left(
        X^{i}\neq \vx^{i}
        \,\middle|\,
        Z^{-i}=\vz^{-i},T=t
    \right)
    &\le
    \frac{
        (N-1)
        (1+\rho_t)^{M^i(\vx,\vz)-\Delta}
    }{
        (1+\rho_t)^{M^i(\vx,\vz)}
    }
    \notag\\
    &=
    (N-1)(1+\rho_t)^{-\Delta}.
    \label{eq:finite_wrong_target_bound}
\end{align}

Combining
\eqref{eq:finite_fisher_variance},
\eqref{eq:finite_fisher_wrong_target}, and
\eqref{eq:finite_wrong_target_bound} yields
\begin{align}
    \mathcal I_{\alpha_t}^{i}(\vz)
    &\le
    (N-1)(L-1)^2
    \left(
        \frac{\mathrm{d}}{\mathrm{d}\alpha_t}\log(1+\rho_t)
    \right)^2
    (1+\rho_t)^{-\Delta}.
    \label{eq:finite_fisher_rho_bound}
\end{align}

Using
\[
\rho_t=\frac{K\alpha_t}{1-\alpha_t},
\]
we have
\begin{align}
    1+\rho_t
    &=
    \frac{1+(K-1)\alpha_t}{1-\alpha_t},
    \label{eq:finite_alpha_one_plus_rho}\\
    \frac{\mathrm{d}}{\mathrm{d}\alpha_t}\log(1+\rho_t)
    &=
    \frac{K}{
        (1-\alpha_t)
        \left(1+(K-1)\alpha_t\right)
    }.
    \label{eq:finite_alpha_eta_derivative}
\end{align}
Substituting
\eqref{eq:finite_alpha_one_plus_rho} and
\eqref{eq:finite_alpha_eta_derivative} into
\eqref{eq:finite_fisher_rho_bound} gives
\begin{align}
    \mathcal I_{\alpha_t}^{i}(\vz)
    &\le
    (N-1)(L-1)^2 K^2
    \frac{
        (1-\alpha_t)^{\Delta-2}
    }{
        \left(1+(K-1)\alpha_t\right)^{\Delta+2}
    }.
    \label{eq:finite_fisher_alpha_bound}
\end{align}

It remains to establish the monotonicity of the bound. Define
\[
g(\alpha)
\coloneqq
\frac{
    (1-\alpha)^{\Delta-2}
}{
    \left(1+(K-1)\alpha\right)^{\Delta+2}
}.
\]
For $\Delta\ge2$ and $\alpha\in(0,1)$,
\begin{align*}
    \frac{\mathrm{d}}{\mathrm{d}\alpha}\log g(\alpha)
    &=
    -\frac{\Delta-2}{1-\alpha}
    -
    \frac{(\Delta+2)(K-1)}
    {1+(K-1)\alpha}
    <0.
\end{align*}
Hence, $g(\alpha)$ is strictly decreasing in $\alpha$. Since
$\alpha_t$ is strictly decreasing in $t$, $g(\alpha_t)$ is strictly
increasing in $t$. Therefore, the right-hand side of
\eqref{eq:finite_fisher_alpha_bound} is strictly increasing in
$t\in(0,1)$ for $\Delta\ge2$, which proves the claim.
\end{proof}

\subsection{Proof of \eqref{eq:fisher_upper_loose_bound}}

\begin{proof}
By Theorem~\ref{thm:finitefisher},
\begin{align*}
    \mathcal I_{\alpha_t}^{i}(\vz)
    &\le
    (N-1)(L-1)^2 K^2
    \frac{
        (1-\alpha_t)^{\Delta-2}
    }{
        \left(1+(K-1)\alpha_t\right)^{\Delta+2}
    }.
\end{align*}
For $\Delta\ge2$, we have
\[
(1-\alpha_t)^{\Delta-2}\le1
\]
for every $\alpha_t\in(0,1)$. Moreover, suppose
\[
\alpha_t\ge K^{-\beta}.
\]
Then
\begin{align*}
    1+(K-1)\alpha_t
    &\ge
    1+(K-1)K^{-\beta}
    \notag\\
    &=
    K^{1-\beta}+1-K^{-\beta}
    \notag\\
    &\ge
    K^{1-\beta}.
\end{align*}
Consequently,
\begin{align*}
    \mathcal I_{\alpha_t}^{i}(\vz)
    &\le
    (N-1)(L-1)^2
    K^{2-(1-\beta)(\Delta+2)}
    \notag\\
    &=
    (N-1)(L-1)^2
    K^{-\Delta+\beta(\Delta+2)}.
\end{align*}
Taking the supremum over all $t$ satisfying
$\alpha_t\ge K^{-\beta}$ gives
\begin{align*}
    \sup_{t:\,\alpha_t\ge K^{-\beta}}
    \mathcal I_{\alpha_t}^{i}(\vz)
    &\le
    (N-1)(L-1)^2
    K^{-\Delta+\beta(\Delta+2)}.
\end{align*}
The exponent is negative whenever
\[
0<\beta<\frac{\Delta}{\Delta+2}.
\]
Thus, for any
$0<\beta<\frac{\Delta}{\Delta+2}$ and $\Delta\ge2$,
the Fisher sensitivity is polynomially suppressed throughout the region
$\alpha_t\ge K^{-\beta}$.

Choosing $\beta=1/4$, which is valid for every $\Delta\ge2$, gives
\begin{align*}
    \sup_{t:\,\alpha_t\ge K^{-1/4}}
    \mathcal I_{\alpha_t}^{i}(\vz)
    &\le
    (N-1)(L-1)^2
    K^{-(3\Delta-2)/4}.
\end{align*}

This provides a useful intuition for a general schedule $\alpha_t$. If $\vz$ is $\Delta$-separable with $\Delta\ge2$, its Fisher sensitivity is suppressed by a negative power of $K$ whenever the expected clean-token fraction exceeds $K^{-1/4}$. Thus, irrespective of the particular scheduler, a $\Delta$-separable $\vz$ has low empirical-optimal Fisher sensitivity with respect to the noise coordinate throughout this regime.
Finally, under the linear schedule $\alpha_t=1-t$,
the condition $\alpha_t\ge K^{-\beta}$ is equivalent to
\[
0<t\le1-K^{-\beta},
\]
and $\mathcal I_{\alpha_t}^{i}(\vz)=\mathcal I_t^{i}(\vz)$.
Therefore,
\begin{align*}
    \sup_{0<t\le1-K^{-1/4}}
    \mathcal I_t^{i}(\vz)
    &\le
    NL^2
    K^{-(3\Delta-2)/4},
\end{align*}
where we intentionally upper-bound
$(N-1)(L-1)^2<NL^2$ to give a simpler and more intuitive bound in the
main paper.
\end{proof}

\subsection{Discussion on the Finite Data Setting}\label{sec:assumption}

In Section~\ref{sec:analysis}, we specialize the data distribution to the finite empirical distribution $\real$ under Assumption~\ref{asm:finite}. 
Here, we verify that this finite-data formulation is fully consistent with the theoretical results in previous works and in Section~\ref{sec:time_agnostic_udm}. For example, deriving the continuous-time UDM NELBO from its discrete-time counterpart could in principle rely on assumptions on the data distribution, such as full support; here, we explicitly verify that the finite empirical distribution does not invalidate these results.
Specifically, we show that (i) the UDM NELBO remains a valid variational objective under $\real$, (ii) its unrestricted optimum remains the uniform-channel leave-one-out posterior characterized in Proposition~\ref{prop:optimum}, and consequently, (iii) Theorem~\ref{thm:exponential_reliability} and Proposition~\ref{prop:exact_temporal_sensitivity} continue to hold under the finite empirical distribution.

\paragraph{Validity of the UDM NELBO under finite data.}
We first verify that the UDM NELBO remains a valid variational objective
under the finite empirical distribution in Assumption~\ref{asm:finite}.
The derivation of the UDM NELBO~\citep{schiff2024simple} is pointwise in
the clean sequence $\vx$ and does not require any particular form of the
outer data distribution. In particular, for a finite discretization with
$T$ diffusion steps, define $t(i)=\frac{i+1}{T+1}$, $s(i)=\frac{i}{T+1}$, for $i=0,\ldots,T$. Then, for every $\vx\in\calX^L$,
\begin{align*}
-\log p_\theta(\vx)
\le\;
\E_q\Bigg[
    -\log p_\theta\!\left(\vx\mid\vz_{t(0)}\right)
    &+\sum_{i=1}^{T}
    \mathcal{D}_{\mathrm{KL}}\!\left[
        q\!\left(
            \vz_{s(i)}
            \mid
            \vz_{t(i)},\vx
        \right)
        \,\middle\|\,
        p_\theta\!\left(
            \vz_{s(i)}
            \mid
            \vz_{t(i)}
        \right)
    \right]
\Bigg]
\nonumber\\
&\quad+
\mathcal{D}_{\mathrm{KL}}\!\left[
    q\!\left(
        \vz_{t(T)}
        \mid
        \vx
    \right)
    \,\middle\|\,
    p_\theta\!\left(
        \vz_{t(T)}
    \right)
\right].
\end{align*}

Crucially, this bound holds pointwise for every clean sequence
$\vx\in\calX^L$. Its derivation does not depend on the support,
cardinality, or any other property of the data distribution from which
$\vx$ is drawn. 
Therefore, the discrete-time NELBO remains valid when the data distribution is replaced by the finite empirical distribution $\real$ in Assumption~\ref{asm:finite}. We now analyze whether the continuous-time limit of the discrete-time NELBO derived by \citet{schiff2024simple} holds in finite-data setting. Their derivation takes $T\to\infty$ for each fixed clean sequence $\vx$. In particular, the prior loss vanishes because
$\alpha_{t(T)}=0$ implies
\[
q\!\left(\vz_{t(T)}^{i}\mid\vx\right)=\mathrm{Cat}(\cdot ; K^{-1}\vone_K),
\]
independently of $\vx$, so that setting
$p_\theta(\vz_{t(T)}^i)=K^{-1}\vone_K$ gives zero prior KL. Likewise, as
$t(0)=1/(T+1)\to0$ and $\alpha_{t(0)}\to1$,
\[
q\!\left(\vz_{t(0)}^{i}\mid\vx\right)
\longrightarrow
\mathrm{Cat}(\cdot;\vx^i),
\]
and the endpoint denoising parameterization copies its input, making the
reconstruction loss vanish. Neither argument depends on the support of
the data distribution. Thus, the remaining diffusion term converges pointwise in $\vx$ to
\begin{align*}
\Ls_{\mathrm{UDM}}(\vx;\theta)
=
\int_0^1
\E_{\vz_t\sim q_t(\cdot\mid\vx)}
\sum_{i=1}^{L}
\frac{\alpha_t'}{K\alpha_t}
\Bigg[
&
\frac{K}{\bar{x}_{\ell_i}^{i}}
-\frac{K}{\bar{x}_{\theta,\ell_i}^{i}}-
\sum_{j\neq \ell_i}
\frac{\bar{x}_{j}^{i}}{\bar{x}_{\ell_i}^{i}}
\log
\frac{
\bar{x}_{\theta,\ell_i}^{i}\bar{x}_{j}^{i}
}{
\bar{x}_{\theta,j}^{i}\bar{x}_{\ell_i}^{i}
}
\Bigg]\mathrm{d}t ,
\end{align*}
where
$\bar{\vnu}^{i}=K\alpha_t\vnu^{i}+(1-\alpha_t)\vone_K$
and $\ell_i$ denotes the non-zero dimension of $\vz_t^{i}$. Taking expectation of $\vx$ then gives:
\begin{align*}
\Ls_{\mathrm{UDM}}(\theta)
=\int_0^1\E_{\vx,\vz_t}\sum_{i=1}^L
\frac{\alpha_t'}{K\alpha_t}
\left[
\frac{K}{\bar{x}_{\ell_i}^{i}}
-\frac{K}{\bar{x}_{\theta,\ell_i}^{i}}
-\sum_{j\neq \ell_i}
\frac{\bar{x}_{j}^{i}}{\bar{x}_{\ell_i}^{i}}
\log\frac{\bar{x}_{\theta,\ell_i}^{i}\bar{x}_{j}^{i}}
{\bar{x}_{\theta,j}^{i}\bar{x}_{\ell_i}^{i}}
\right]\mathrm{d}t,
\end{align*}
where $\vx$ comes from $p_{\mathrm{data}}$ or $\real$.Hence, the continuous-time UDM NELBO is likewise well defined for each
individual clean sequence, irrespective of the support of the data
distribution.

\paragraph{Validity of the leave-one-out optimum under finite data.}
We next verify that the leave-one-out characterization of the UDM
NELBO optimum remains valid under Assumption~\ref{asm:finite}. We refer the readers to Appendix~A.1 of \citet{gourevitch2026uniformdiffusionmodelsrevisited} for a full understanding.
The proof of their optimum results applies directly when $p_{\mathrm{data}}$ is replaced by $\real$. Their argument relies
only on Bayes' rule, the Chapman--Kolmogorov identity, factorization of
the bridge across coordinates, and the Bayes-compatible simplex
extension of the plug-in bridge. None of these steps imposes any
assumption on the cardinality or support structure of the clean-data
distribution.

\paragraph{Validity of the theoretical results in Section~\ref{sec:time_agnostic_udm}.}
Having established that the UDM NELBO and its leave-one-out optimum
remain valid under Assumption~\ref{asm:finite}, the remaining results in
Section~\ref{sec:time_agnostic_udm} follow without any additional
assumption on the data distribution. Proposition~\ref{prop:optimum} is
obtained from the leave-one-out posterior using only Bayes' rule,
factorization of the uniform forward channel across coordinates, and a
finite subset expansion. These arguments apply identically after
replacing $p_{\mathrm{data}}$ by $\real$.

Likewise, Theorem~\ref{thm:exponential_reliability} only groups the
probability mass of the data distribution according to the matched-context
count $M^i$ and uses the fact that sequences with the same match count
have the same uniform-channel likelihood. Proposition~
\ref{prop:exact_temporal_sensitivity} then follows by differentiating the
resulting finite normalized mixture weights. None of these derivations
requires any assumption on the cardinality or support structure of
$p_{\mathrm{data}}$. Therefore, Proposition~\ref{prop:optimum},
Theorem~\ref{thm:exponential_reliability}, and Proposition~
\ref{prop:exact_temporal_sensitivity} all remain valid under the finite
empirical distribution in Assumption~\ref{asm:finite}.

\section{Experimental Details}\label{sec:experimental_details}
\subsection{Additional Experimental Results}\label{sec:gen_ppl}
\begin{figure}[t]
    \centering

    \begin{subfigure}[t]{0.49\linewidth}
        \centering
        \includegraphics[width=\linewidth]
        {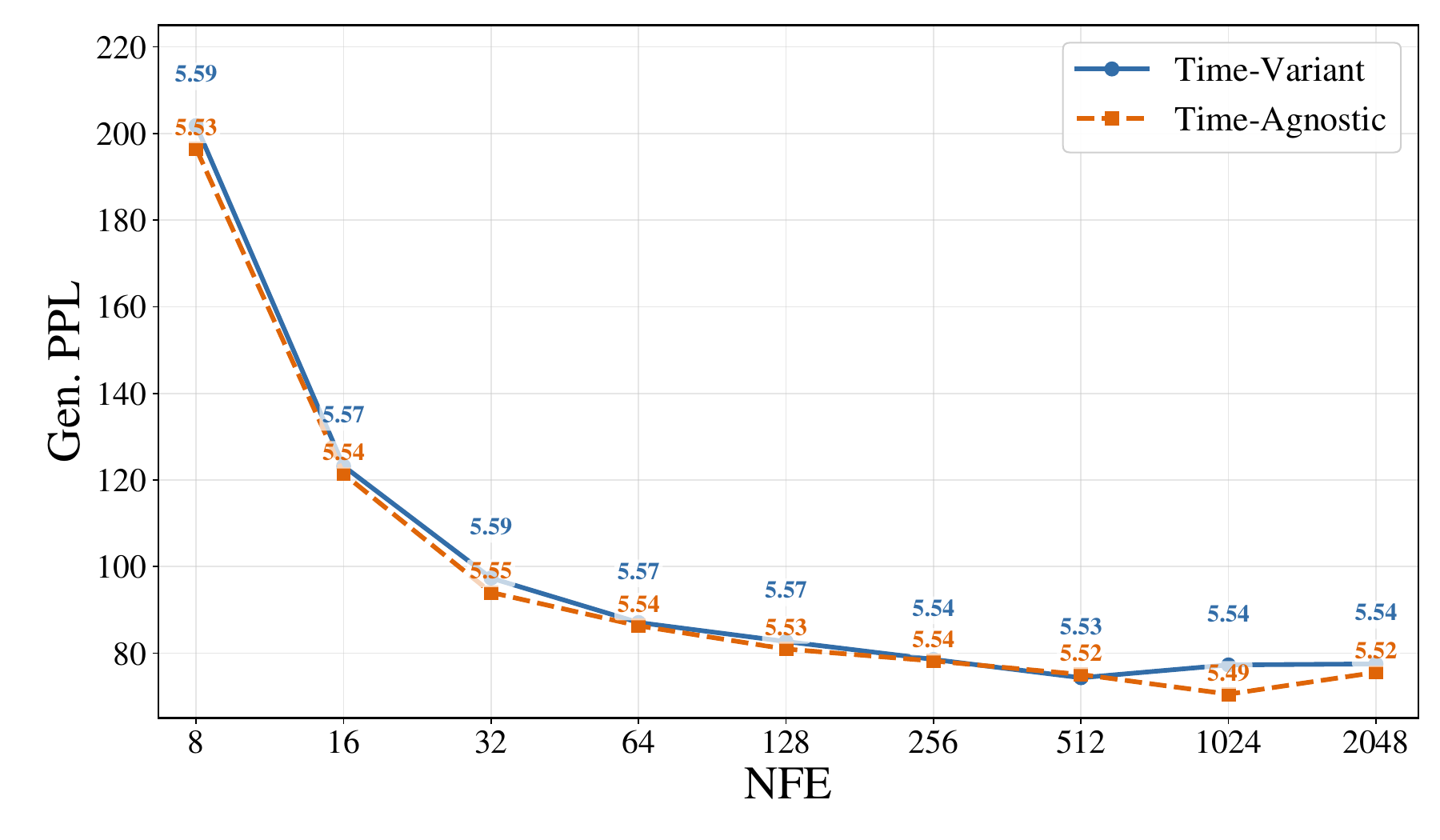}
        \caption{\textsc{Duality}}
        \label{fig:genppl_nfe_duality}
    \end{subfigure}
    \hfill
    \begin{subfigure}[t]{0.49\linewidth}
        \centering
        \includegraphics[width=\linewidth]
        {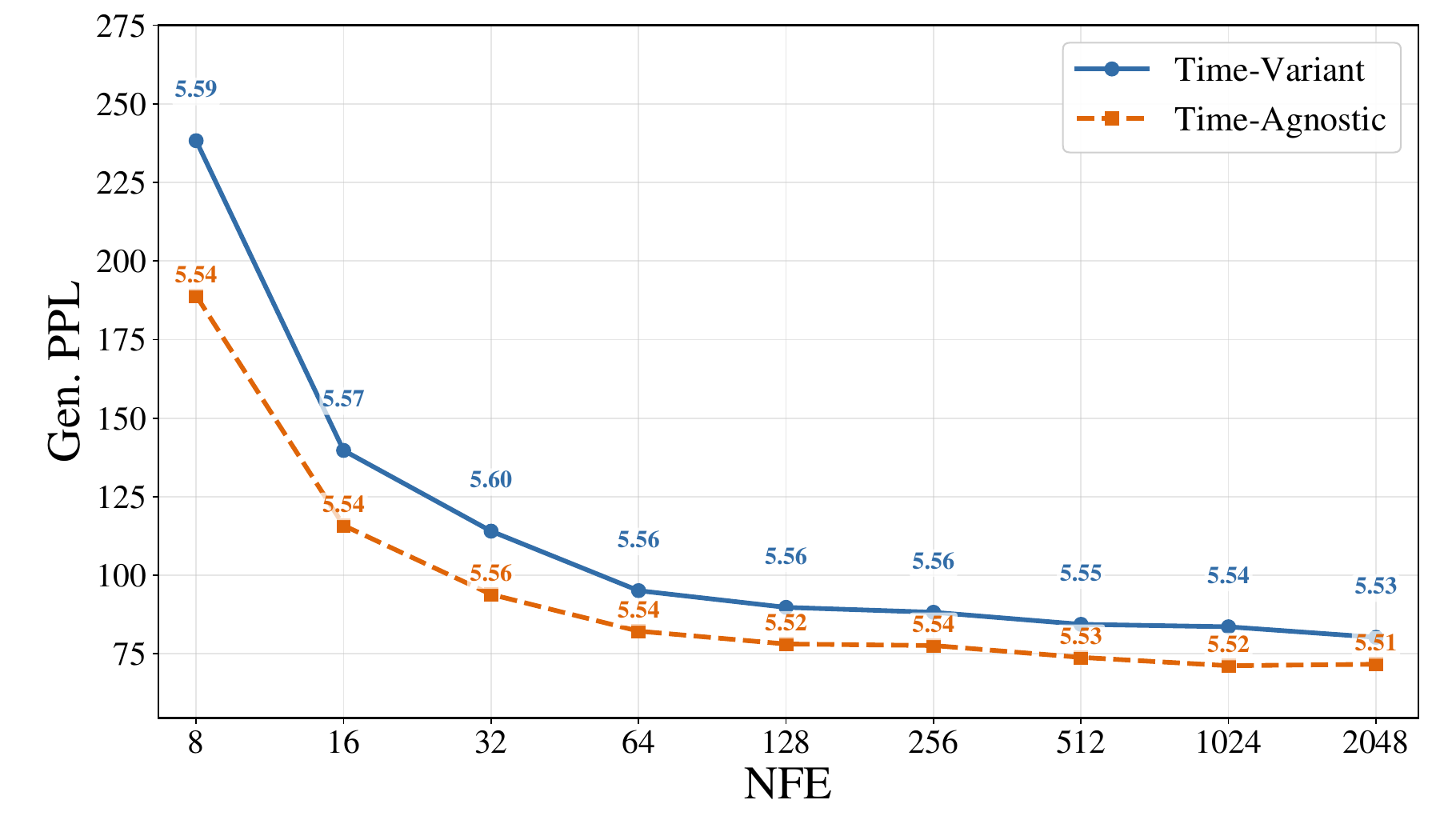}
        \caption{\textsc{LOO+CE}}
        \label{fig:genppl_nfe_reudm}
    \end{subfigure}

    \caption{
        Sampling performance of time-variant and time-agnostic UDMs across
        different numbers of function evaluations (NFEs).
        We compare generative perplexity as the sampling budget varies under
        \textbf{(a)} \textsc{Duality} and \textbf{(b)} \textsc{LOO+CE};
        the annotated values report the entropy of generated samples.
    }
    \label{fig:genppl_vs_nfe}
\end{figure}

\begin{figure}
    \centering
    \includegraphics[width=\linewidth]{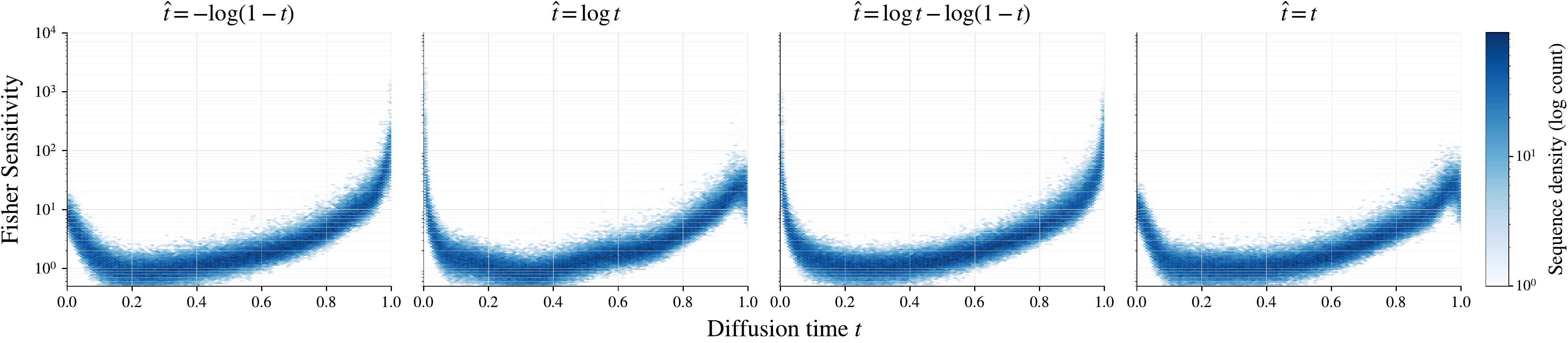}
    \caption{Fisher sensitivity under different time-input reparameterizations, where $\vz$ sampling time and Fisher sensitivity evaluation time are equal. All sensitivities are measured with respect to the original diffusion time $t$, not reparametrized time $\hat t$ for model input.}
    \label{fig:learned_time_dependence_appendix}
\end{figure}

We provide generative perplexity results for \textsc{Duality} and \textsc{LOO+CE} trained on OWT, which were omitted from the main paper.
For each NFE, we generate 256 samples.
Following \citet{zheng2024maskeddiffusionmodelssecretly} and \textsc{Duality}~\citep{sahoo2025diffusionduality}, we use float64 precision for sampling, as low-precision sampling can lead to reduced diversity and misleading generative perplexity in discrete diffusion models.
As shown in Figure~\ref{fig:genppl_vs_nfe} (\textit{Left}), time-agnostic and time-variant \textsc{Duality} achieve nearly identical generative perplexity across a wide range of NFEs.
Their sample entropies are also closely matched, indicating similar sampling quality and diversity. On the other hand, as shown in Figure~\ref{fig:genppl_vs_nfe} (\textit{Right}), \textsc{LOO+CE} exhibits similar sample entropy for the time-agnostic and time-variant models, while the time-agnostic model achieves lower generative perplexity.

\subsection{Experimental Settings}\label{sec:exp_settings}
\noindent\textbf{Settings on training experiments.} We use exactly the same experimental settings as the original UDM~\citep{schiff2024simple} and Diffusion Duality~\citep{sahoo2025diffusionduality}.
We detokenize the One Billion Words dataset following \citet{lou2024discrete,sahoo2024simple}, whose official code can be found. We tokenize LM1B using the \textsc{bert-base-uncased} tokenizer, consistent with \citet{he2022diffusionbert}. 
We then concatenate and pack the sequences to a fixed length of 128 \citep{raffel2020exploring}. 
For OWT, we use the GPT-2 tokenizer and similarly concatenate and pack sequences to length 1{,}024; during packing, we insert an \texttt{eos} token between consecutive documents. 
As OWT does not provide an official validation split, we reserve the last 100k documents for validation.
We parameterize the diffusion backbone of all models with the modified diffusion transformer
architecture \citep{peebles2023scalable} from \citet{lou2024discrete,sahoo2024simple}. For time-variant models, we use 12 layers, a hidden dimension of
768, 12 attention heads.
For the diffusion backbone, we use the AdamW optimizer with a batch size of 512, constant learning rate warmup from 0 to a learning rate of 3e-4 for 2,500 steps, following prior works~\citep{sahoo2024simple,lou2024discrete,sahoo2025diffusionduality}. We use a dropout rate of 0.1. See Appendix~\ref{sec:ta-architecture} for time-agnostic and hybrid models.

\noindent\textbf{Experimental setting for the Hamming-distance and $\Delta$ analyses (Figure~\ref{fig:data_separation})} The minimum hamming distance histogram is obtained by exhaustively comparing all $8{,}730{,}966$ OWT training rows of length $L=1024$.
For each row, we compute its exact minimum Hamming distance as $L$ minus the maximum position-wise match count over all other rows, excluding its own row index.
Special tokens are treated like ordinary tokens throughout.

For each $t\in\{0.05,0.10,\ldots,0.95,0.99\}$, we independently sample $2{,}000$ source rows uniformly with replacement and corrupt each token by copying it with probability $1-t$ or drawing uniformly from the full $50{,}258$-token vocabulary, including the added padding token, with probability $t$.
For each pair, we independently sample $i\sim\operatorname{Unif}([L])$ and compute $\Delta_i=M^i(\vx,\vz_t)-\max_{\vx':\,\vx'^i\neq\vx^i}M^i(\vx',\vz_t)$ by scanning the entire training set, where $M^i$ counts matches only outside $i$.
The probability mass at each time is estimated from the empirical frequencies of these $2{,}000$ margins.

Both measurements use exact exhaustive GPU scans, without approximate nearest-neighbor search.
We store the entire token matrix losslessly in 16-bit integers (approximately $17.9$\,GB) on each H100 GPU and compute integer match counts across training rows. 

\noindent\textbf{Experimental setting for the learned time-sensitivity analysis (Figure~\ref{fig:learned_time_dependence} (\textit{left})).}
We use the public checkpoint both for \textsc{Duality} and \textsc{LOO+CE} trained on OWT.
We uniformly sample $1000$ held-out OWT sequences of length $L=1024$ without replacement.
For each $t_{\mathrm{corruption}}\in\{0.01,0.50,0.99\}$, we independently draw one corrupted sequence per clean sequence from $q_{t_{\mathrm{corruption}}}$ and reuse identical inputs for both models.
Each corrupted input $\vz$ is then held fixed while the evaluation time varies over $t\in\{0.01,0.02,\ldots,0.99\}$.
We apply \eqref{eq:fisher_sensitivity} to the raw network softmax prediction $\vx_\theta^i(\vz,t)$ in place of the oracle, before any analytic local-token correction.
Derivatives are computed using automatic differentiation with respect to raw $t$.
Each curve reports the arithmetic mean over all valid token positions, including unchanged tokens, followed by the mean over the $1000$ sequences. In particular, for \textsc{LOO+CE}, we measure Fisher sensitivity with respect to $\vx_\theta$, the LOO denoiser actually used for generation and in the plug-in reverse kernel, rather than $\vmu_\theta^{\mathrm{Posterior}}$, which is introduced through the LOO reparameterization in \eqref{eq:loo_parametrization}.

\noindent\textbf{Perplexity measure.}
In all experiments, we estimate validation/test perplexity by exponentiating the
UDM NELBO. That is, we report perplexity that corresponds to an upper bound on the true perplexity, following the standard convention in most discrete diffusion language modeling works~\citep{sahoo2024simple,sahoo2025diffusionduality,schiff2024simple,gourevitch2026uniformdiffusionmodelsrevisited,shi2025simplified}. Likewise, the time-resolved NLL plots report the expected token-normalized NELBO integrand at time $t$.

\subsection{Details on Time-agnostic Models}\label{sec:ta-architecture}
\noindent\textbf{Time-agnostic UDMs.}
Across all frameworks---including the conventional UDM~\citep{schiff2024simple}, Diffusion Duality~\citep{sahoo2025diffusionduality},
and the leave-one-out denoiser trained with cross-entropy~\citep{gourevitch2026uniformdiffusionmodelsrevisited}---the only
modification we make is to render the model time-agnostic.
All other framework-specific components, including the loss function,
forward process, noise schedule, and curriculum learning procedure, are
kept exactly as originally proposed.
More specifically, we directly build upon the original codebases and leave
their framework-specific implementations unchanged, except for removing
time conditioning from the model.
We refer the reader to Appendix~\ref{sec:detailed_background} for a detailed
description of these frameworks. For experiments on LM1B, we compared 1) time-variant model with 12 Transformer layers, 2) time-agnostic model with 12 Transformer layers, 3) time-agnostic model with 13 Transformer layers that matches parameter count of 1), and 4) hybrid model. For experiments on OWT, we compared the time-variant model with 12 Transformer layers and the time-agnostic model with 13 Transformer layers that matches the parameter count of the time-variant counterpart. All architectures are explained in detail below.

\noindent\textbf{Time-agnostic architecture.}
We use the same Transformer backbone for the time-conditioned and
time-agnostic models, differing only in how the normalization and residual
modulation parameters are specified.
For the time-conditioned model, we follow the backbone of
\citet{sahoo2025diffusionduality,schiff2024simple}, where each Transformer
block uses the adaLN-Zero parameterization: a timestep embedding produces
time-dependent scale, shift, and residual-gating parameters. For time-agnostic models, we simply replaced time-dependent parameters with single learnable parameters.

More precisely, let $\mathbf{c}(t)\in\mathbb{R}^{d_c}$ denote the timestep
embedding, where $d_c$ is the conditioning dimension, and let $d$ denote
the Transformer hidden dimension.
For the $\ell$-th Transformer block, the modulation parameters are generated as
\[
\begin{aligned}
\bigl(
&\boldsymbol{\beta}_{\ell,\mathrm{attn}}(t),
\mathbf{s}_{\ell,\mathrm{attn}}(t),
\mathbf{g}_{\ell,\mathrm{attn}}(t),
\boldsymbol{\beta}_{\ell,\mathrm{mlp}}(t),
\mathbf{s}_{\ell,\mathrm{mlp}}(t),
\mathbf{g}_{\ell,\mathrm{mlp}}(t)
\bigr)
=
\mathbf{W}_{\ell}\mathbf{c}(t)+\mathbf{b}_{\ell},
\end{aligned}
\]
where
$\mathbf{W}_{\ell}\in\mathbb{R}^{6d\times d_c}$,
$\mathbf{b}_{\ell}\in\mathbb{R}^{6d}$, and each modulation vector lies in
$\mathbb{R}^{d}$.
Here, $\boldsymbol{\beta}$ and $\mathbf{s}$ denote the channel-wise shift
and scale modulation of LayerNorm, respectively, while $\mathbf{g}$ denotes
the channel-wise residual gate.

Let $\mathbf{h}_{\ell}\in\mathbb{R}^{L\times d}$ denote the sequence of
hidden representations entering the $\ell$-th block.
Ignoring dropout for notational simplicity, the block is
\[
\begin{aligned}
\widetilde{\mathbf{h}}_{\ell}
&=
\mathbf{h}_{\ell}
+
\mathbf{g}_{\ell,\mathrm{attn}}(t)
\odot
\operatorname{Attn}_{\ell}
\left(
\bigl(\vone_d+\mathbf{s}_{\ell,\mathrm{attn}}(t)\bigr)
\odot
\operatorname{LN}_{\mathbf{w}_{\ell,\mathrm{attn}}}
(\mathbf{h}_{\ell})
+
\boldsymbol{\beta}_{\ell,\mathrm{attn}}(t)
\right),\\
\mathbf{h}_{\ell+1}
&=
\widetilde{\mathbf{h}}_{\ell}
+
\mathbf{g}_{\ell,\mathrm{mlp}}(t)
\odot
\operatorname{MLP}_{\ell}
\left(
\bigl(\vone_d+\mathbf{s}_{\ell,\mathrm{mlp}}(t)\bigr)
\odot
\operatorname{LN}_{\mathbf{w}_{\ell,\mathrm{mlp}}}
(\widetilde{\mathbf{h}}_{\ell})
+
\boldsymbol{\beta}_{\ell,\mathrm{mlp}}(t)
\right).
\end{aligned}
\]
Here,
$\mathbf{w}_{\ell,\mathrm{attn}},
\mathbf{w}_{\ell,\mathrm{mlp}}\in\mathbb{R}^{d}$
are the learned channel-wise LayerNorm scales.
All element-wise operations above act along the hidden dimension, with the
$d$-dimensional modulation vectors broadcast across sequence positions.
Thus, time enters the Transformer backbone only through the scale, shift,
and residual-gating parameters of each block.

For the time-agnostic model, we remove the timestep embedding
$\mathbf{c}(t)$ entirely and directly learn a time-independent set of
modulation parameters for each block:
\[
\bigl(
\boldsymbol{\beta}_{\ell,\mathrm{attn}},
\mathbf{s}_{\ell,\mathrm{attn}},
\mathbf{g}_{\ell,\mathrm{attn}},
\boldsymbol{\beta}_{\ell,\mathrm{mlp}},
\mathbf{s}_{\ell,\mathrm{mlp}},
\mathbf{g}_{\ell,\mathrm{mlp}}
\bigr)
\in\mathbb{R}^{6d},
\]
where each of the six vectors belongs to $\mathbb{R}^{d}$.The resulting static normalization modulation is equivalent to a standard affine LayerNorm. More precisely, for a token representation $\mathbf{h}\in\mathbb{R}^{d}$, define 
\[
\operatorname{LN}_{0}(\mathbf{h})
=
\frac{
\mathbf{h}-\mu(\mathbf{h})\vone_d
}{
\sqrt{\operatorname{Var}(\mathbf{h})+\epsilon}
}, \quad \operatorname{LN}_{\mathbf{w}}(\mathbf{h})
=
\mathbf{w}\odot\operatorname{LN}_{0}(\mathbf{h}).
\]
Then, for time-independent
$\mathbf{s},\boldsymbol{\beta}\in\mathbb{R}^{d}$,
\[
\begin{aligned}
(\vone_d+\mathbf{s})
\odot
\operatorname{LN}_{\mathbf{w}}(\mathbf{h})
+
\boldsymbol{\beta}
=
\bigl(
\mathbf{w}\odot(\vone_d+\mathbf{s})
\bigr)
\odot
\operatorname{LN}_{0}(\mathbf{h})
+
\boldsymbol{\beta}=
\boldsymbol{\gamma}
\odot
\operatorname{LN}_{0}(\mathbf{h})
+
\boldsymbol{\beta},
\end{aligned}
\]
where $\boldsymbol{\gamma}
=
\mathbf{w}\odot(\vone_d+\mathbf{s})
\in\mathbb{R}^{d}$.
The last expression is precisely standard affine LayerNorm with learned
channel-wise scale $\boldsymbol{\gamma}$ and shift $\boldsymbol{\beta}$. Thus, the time-agnostic block is equivalent to
\[
\begin{aligned}
\widetilde{\mathbf{h}}_{\ell}
&=
\mathbf{h}_{\ell}
+
\mathbf{g}_{\ell,\mathrm{attn}}
\odot
\operatorname{Attn}_{\ell}
\left(
\operatorname{LN}^{\mathrm{attn}}_{\ell}(\mathbf{h}_{\ell})
\right),\\
\mathbf{h}_{\ell+1}
&=
\widetilde{\mathbf{h}}_{\ell}
+
\mathbf{g}_{\ell,\mathrm{mlp}}
\odot
\operatorname{MLP}_{\ell}
\left(
\operatorname{LN}^{\mathrm{mlp}}_{\ell}
(\widetilde{\mathbf{h}}_{\ell})
\right),
\end{aligned}
\]
where
$\operatorname{LN}^{\mathrm{attn}}_{\ell}$ and
$\operatorname{LN}^{\mathrm{mlp}}_{\ell}$ are standard learned affine
LayerNorms with channel-wise scale and shift in $\mathbb{R}^{d}$, and
$\mathbf{g}_{\ell,\mathrm{attn}},
\mathbf{g}_{\ell,\mathrm{mlp}}\in\mathbb{R}^{d}$
are learned, time-independent residual gates initialized to zero.

We apply the same replacement to the time-conditioned modulation in the final output layer.

\noindent\textbf{Hybrid architecture.}
The hybrid model combines the time-conditioned and time-agnostic
parameterizations above, using explicit time conditioning only in the high-noise
regime. Specifically, for each Transformer block, we retain both the original
time-modulation layer and a separate set of directly learned, time-independent
modulation parameters. The six modulation vectors are given by
\[
\bigl(\boldsymbol{\beta}_{\ell,r},\mathbf{s}_{\ell,r},\mathbf{g}_{\ell,r}\bigr)_{r\in\{\mathrm{attn},\mathrm{mlp}\}}
=
\begin{cases}
\mathbf{W}_{\ell}\mathbf{c}(t)+\mathbf{b}_{\ell}, & t>0.8,\\
\bigl(\boldsymbol{\beta}^{0}_{\ell,r},\mathbf{s}^{0}_{\ell,r},\mathbf{g}^{0}_{\ell,r}\bigr)_{r\in\{\mathrm{attn},\mathrm{mlp}\}}, & t\le0.8.
\end{cases}
\]
where the six vectors in the second branch are learned directly and do not
depend on time. Thus, the model is identical to the time-conditioned
architecture for $t>0.8$ and to the time-agnostic architecture for
$t\leq0.8$, with all other components left unchanged.

\end{document}

%% file: math_commands.tex
\usepackage{amsmath,amsfonts,bm}

\def\eqref#1{equation~\ref{#1}}

\def\1{\bm{1}}

\def\vone{{\bm{1}}}
\def\vmu{{\bm{\mu}}}

\def\ve{{\bm{e}}}

\def\vp{{\bm{p}}}

\def\vw{{\bm{w}}}
\def\vx{{\bm{x}}}

\def\vz{{\bm{z}}}

\def\mI{{\bm{I}}}

\def\mP{{\bm{P}}}
\def\mQ{{\bm{Q}}}

\DeclareMathAlphabet{\mathsfit}{\encodingdefault}{\sfdefault}{m}{sl}
\SetMathAlphabet{\mathsfit}{bold}{\encodingdefault}{\sfdefault}{bx}{n}

\newcommand{\E}{\mathbb{E}}
\newcommand{\Ls}{\mathcal{L}}

\newcommand{\softmax}{\mathrm{softmax}}

\DeclareMathOperator*{\argmax}{arg\,max}

%% file: def.tex
\newcommand{\calN}{{\mathcal{N}}}

\newcommand{\calT}{{\mathcal{T}}}

\newcommand{\calX}{{\mathcal{X}}}

\newcommand{\matrixb}{\left[ \begin{array}}
\newcommand{\matrixe}{\end{array} \right]}

\usepackage{xspace}
\makeatletter
\DeclareRobustCommand\onedot{\futurelet\@let@token\@onedot}
\def\@onedot{\ifx\@let@token.\else.\null\fi\xspace}

\def\ie{\emph{i.e}\onedot}

\makeatother

\newcommand{\Cref}[1]{Chap.~\ref{#1}}

\renewcommand{\paragraph}[1]{\vspace{1mm}\noindent\textbf{#1}\,\,\,}

\usepackage{enumitem}
\setlist[itemize]{align=parleft,left=0pt}

%% file: table.tex

\begingroup

\newcommand{\better}[1]{\underline{#1}}
\newcommand{\tableNA}{\textemdash}

\begin{table}[t]
  \centering
  \caption{
    Validation perplexity (PPL) across several configurations. Lower is better.
    Hybrid uses the time-modulation layer only for $t>0.8$.
    PPLs lower than the time-variant baseline are underlined.
  }
  \label{tab:best-validation-ppl}

  \renewcommand{\arraystretch}{1.20}
  \setlength{\tabcolsep}{12pt}

  \begin{adjustbox}{max width=0.98\textwidth,center}
  \begin{tabular}{@{}llcccc@{}}

    \toprule

    Dataset &
    Objective &
    Variant (139M) &
    Hybrid (139M) &
    Agnostic (131M) &
    Agnostic (139M) \\

    \midrule

    \multirow{2}{*}{LM1B}
      & \textsc{Duality}
      & 29.99
      & \better{29.92}
      & 30.32
      & \better{29.76} \\

      & \textsc{LOO+CE}
      & 30.15
      & \better{29.67}
      & \better{29.18}
      & \better{28.93} \\

    \addlinespace[2pt]

    \multirow{2}{*}{LM1B (Packed)}
      & \textsc{Duality}
      & 33.48
      & \better{33.17}
      & 33.69
      & \better{33.35} \\

      & \textsc{LOO+CE}
      & 33.94
      & \better{33.37}
      & \better{33.24}
      & \better{32.75} \\

    \bottomrule

  \end{tabular}
  \end{adjustbox}
\end{table}
\endgroup

%% file: main.bib
@STRING{JMLR	= "Journal of Machine Learning Research (JMLR)"}

@STRING{NIPS	= "Advances in Neural Information Processing Systems (NeurIPS)"}

@STRING{ICCV	= "IEEE International Conference on Computer Vision (ICCV)"}

@STRING{ICML	= "International Conference on Machine Learning (ICML)"}

@STRING{ICLR	= "International Conference on Learning Representations (ICLR)"}

@STRING{ACL = "Annual Meeting of the Association for Computational Linguistics (ACL)"}

@inproceedings{
    amin2025maskingdiffusionworkscondition,
    title={Why Masking Diffusion Works: Condition on the Jump Schedule for Improved Discrete Diffusion},
    author={Alan Nawzad Amin and Nate Gruver and Andrew Gordon Wilson},
    booktitle=NIPS,
    year={2025},
}

@inproceedings{arriola2025bd3lm,
    author = {Marianne Arriola and Aaron Gokaslan and Justin T. Chiu and Zhihan Yang and Zhixuan Qi and Jiaqi Han and Subham Sekhar Sahoo and Volodymyr Kuleshov},
    title = {Block Diffusion: Interpolating Between Autoregressive and Diffusion Language Models},
    booktitle = ICLR,
    year = {2025}
}

@inproceedings{Austin2021,
  title={Structured denoising diffusion models in discrete state-spaces},
  author={Austin, Jacob and Johnson, Daniel D and Ho, Jonathan and Tarlow, Daniel and Van Den Berg, Rianne},
  booktitle=NIPS,
  year={2021}
}

@inproceedings{campbell2022continuous,
  title={A continuous time framework for discrete denoising models},
  author={Campbell, Andrew and Benton, Joe and De Bortoli, Valentin and Rainforth, Thomas and Deligiannidis, George and Doucet, Arnaud},
  booktitle=NIPS,
  year={2022}
}

@inproceedings{chelba2013lm1b,
  title     = {{One billion word benchmark for measuring progress in statistical language modeling}},
  author    = {Ciprian Chelba and Tomas Mikolov and Mike Schuster and Qi Ge and Thorsten Brants and Phillipp Koehn and Tony Robinson},
  year      = {2014},
  booktitle = {{Interspeech 2014}},
  pages     = {2635--2639},
  doi       = {10.21437/Interspeech.2014-564},
  issn      = {2958-1796},
}

@misc{googledeepmind2026diffusiongemma,
  title={{DiffusionGemma} technical report},
  author={{DiffusionGemma Team}},
  year={2026},
  eprint={2608.00146},
  archivePrefix={arXiv},
  primaryClass={cs.CL},
  url={https://arxiv.org/abs/2608.00146}, 
}

@misc{Gokaslan2019owt,
    title={OpenWebText Corpus},
    author={Gokaslan, Aaron and Cohen, Vanya and Pavlick, Ellie and Tellex, Stefanie},
    howpublished={\url{http://Skylion007.github.io/OpenWebTextCorpus}},
    year={2019}
}

@misc{gourevitch2026uniformdiffusionmodelsrevisited,
      title={Uniform Diffusion Models Revisited: Leave-One-Out Denoiser and Absorbing State Reformulation}, 
      author={Samson Gourevitch and Yazid Janati and Dario Shariatian and Umut Simsekli and Eric Moulines and Eric P. Xing and Alain Durmus},
      year={2026},
      eprint={2605.22765},
      archivePrefix={arXiv},
      primaryClass={cs.LG},
      url={https://arxiv.org/abs/2605.22765}, 
}

@inproceedings{he2022diffusionbert,
  title={{D}iffusion{BERT}: Improving Generative Masked Language Models with Diffusion Models},
  author={He, Zhengfu and Sun, Tianxiang and Tang, Qiong and Wang, Kuanning and Huang, Xuanjing and Qiu, Xipeng},
  booktitle=ACL,
  year={2023}
}

@misc{helbling2026timeitdatageometry,
      title={What Time Is It? How Data Geometry Makes Time Conditioning Optional for Flow Matching}, 
      author={Alec Helbling and Sebastian Gutierrez Hernandez and Benjamin Hoover and Duen Horng Chau and Parikshit Ram},
      year={2026},
      eprint={2605.08344},
      archivePrefix={arXiv},
      primaryClass={cs.LG},
      url={https://arxiv.org/abs/2605.08344}, 
}

@inproceedings{ho2020denoising,
  title={Denoising diffusion probabilistic models},
  author={Ho, Jonathan and Jain, Ajay and Abbeel, Pieter},
  booktitle=NIPS,
  year={2020}
}

@inproceedings{
    hong2025improvingdiscretediffusionunmasking,
    title={Improving Discrete Diffusion Unmasking Policies Beyond Explicit Reference Policies},
    author={Chunsan Hong and Seonho An and Min-Soo Kim and Jong Chul Ye},
    booktitle=ICLR,
    year={2026},
}

@inproceedings{Hoogeboom2021b,
  title={Argmax flows and multinomial diffusion: Learning categorical distributions},
  author={Hoogeboom, Emiel and Nielsen, Didrik and Jaini, Priyank and Forr{\'e}, Patrick and Welling, Max},
  booktitle=NIPS,
  year={2021}
}

@misc{k2horizon2026,
  title  = {Introducing K2 Horizon: Frontier Performance, Radically Open},
  author = {{IFM Team}},
  year   = {2026},
  url    = {https://ifm.ai/blog/k2/},
}

@misc{lai2025principles,
      title={The Principles of Diffusion Models}, 
      author={Chieh-Hsin Lai and Yang Song and Dongjun Kim and Yuki Mitsufuji and Stefano Ermon},
      year={2025},
      eprint={2510.21890},
      archivePrefix={arXiv},
      primaryClass={cs.LG},
      url={https://arxiv.org/abs/2510.21890}, 
}

@misc{lee2025iterref,
      title={Effective Test-Time Scaling of Discrete Diffusion through Iterative Refinement}, 
      author={Sanghyun Lee and Sunwoo Kim and Seungryong Kim and Jongho Park and Dongmin Park},
      year={2025},
      eprint={2511.05562},
      archivePrefix={arXiv},
      primaryClass={cs.LG},
      url={https://arxiv.org/abs/2511.05562}, 
}

@inproceedings{
    liu2025thinkgeneratediscretediffusion,
    title={Think while You Generate: Discrete Diffusion with Planned Denoising},
    author={Sulin Liu and Juno Nam and Andrew Campbell and Hannes Stark and Yilun Xu and Tommi Jaakkola and Rafael Gomez-Bombarelli},
    booktitle=ICLR,
    year={2025},
}

@inproceedings{lou2024discrete,
  title={Discrete Diffusion Modeling by Estimating the Ratios of the Data Distribution},
  author={Aaron Lou and Chenlin Meng and Stefano Ermon},
  booktitle=ICML,
  year={2024}
}

@inproceedings{
    nie2025llada,
    title={Large Language Diffusion Models},
    author={Shen Nie and Fengqi Zhu and Zebin You and Xiaolu Zhang and Jingyang Ou and Jun Hu and Jun Zhou and Yankai Lin and Ji-Rong Wen and Chongxuan Li},
    booktitle=NIPS,
    year={2025},
}

@inproceedings{
    ou2024RADD,
    title={Your Absorbing Discrete Diffusion Secretly Models the Conditional Distributions of Clean Data},
    author={Jingyang Ou and Shen Nie and Kaiwen Xue and Fengqi Zhu and Jiacheng Sun and Zhenguo Li and Chongxuan Li},
    booktitle=ICLR,
    year={2025},
}

@inproceedings{peebles2023scalable,
  title={Scalable diffusion models with transformers},
  author={Peebles, William and Xie, Saining},
  booktitle=ICCV,
  year={2023}
}

@article{raffel2020exploring,
  author  = {Colin Raffel and Noam Shazeer and Adam Roberts and Katherine Lee and Sharan Narang and Michael Matena and Yanqi Zhou and Wei Li and Peter J. Liu},
  title   = {Exploring the Limits of Transfer Learning with a Unified Text-to-Text Transformer},
  journal = JMLR,
  year    = {2020},
  volume  = {21},
  number  = {140},
  pages   = {1--67},
}

@inproceedings{sahoo2024simple,
  title={Simple and effective masked diffusion language models},
  author = {Sahoo, Subham Sekhar and Arriola, Marianne and Schiff, Yair and Gokaslan, Aaron and Marroquin, Edgar and Chiu, Justin T and Rush, Alexander and Kuleshov, Volodymyr},
  booktitle=NIPS,
  year={2024}
}

@inproceedings{sahoo2025diffusionduality,
      title={The Diffusion Duality}, 
      author={Subham Sekhar Sahoo and Justin Deschenaux and Aaron Gokaslan and Guanghan Wang and Justin Chiu and Volodymyr Kuleshov},
      year={2025},
      booktitle=ICML
}

@misc{sahoo2026uno,
  title={Unlocking Lossless Speedups in LLMs via Discrete Diffusion}, 
  author={Subham Sekhar Sahoo and Lingjie Chen and Khiem Pham and Jonathan Geuter and Chaitanya Dwivedi and Varad Pimpalkhute and Yash Akhauri and Alexander Moreno and Mikhail Yurochkin and Zhenting Wang and Mostafa Elhoushi and Nolan Dey and Shane Bergsma and Joel Hestness and John Thickstun and Eric Xing and Zhengzhong Liu},
  year={2026},
  eprint={2609.04010},
  archivePrefix={arXiv},
  primaryClass={cs.LG},
  url={https://arxiv.org/abs/2609.04010}, 
}

@misc{sahraeeardakan2026geometrynoisediffusionmodels,
      title={The Geometry of Noise: Why Diffusion Models Don't Need Noise Conditioning}, 
      author={Mojtaba Sahraee-Ardakan and Mauricio Delbracio and Peyman Milanfar},
      year={2026},
      eprint={2602.18428},
      archivePrefix={arXiv},
      primaryClass={cs.LG},
      url={https://arxiv.org/abs/2602.18428}, 
}

@inproceedings{
    schiff2024simple,
    title={Simple Guidance Mechanisms for Discrete Diffusion Models},
    author={Yair Schiff and Subham Sekhar Sahoo and Hao Phung and Guanghan Wang and Sam Boshar and Hugo Dalla-torre and Bernardo P de Almeida and Alexander M Rush and Thomas Pierrot and Volodymyr Kuleshov},
    booktitle=ICLR,
    year={2025},
}

@inproceedings{
    shi2025simplified,
    title={Simplified and Generalized Masked Diffusion for Discrete Data},
    author={Jiaxin Shi and Kehang Han and Zhe Wang and Arnaud Doucet and Michalis Titsias},
    booktitle=NIPS,
    year={2024},
}

@inproceedings{sohl2015deep,
  title = 	 {Deep Unsupervised Learning using Nonequilibrium Thermodynamics},
  author = 	 {Sohl-Dickstein, Jascha and Weiss, Eric and Maheswaranathan, Niru and Ganguli, Surya},
  booktitle=ICML,
  year={2015},
}

@inproceedings{sun2025noise,
  title = 	 {Is Noise Conditioning Necessary for Denoising Generative Models?},
  author =       {Sun, Qiao and Jiang, Zhicheng and Zhao, Hanhong and He, Kaiming},
  booktitle=ICML,
  year={2025}
}

@inproceedings{
    vonrutte2026scalingbehaviordiscretediffusion,
    title={Scaling Behavior of Discrete Diffusion Language Models},
    author={Dimitri von R{\"u}tte and Janis Fluri and Omead Pooladzandi and Bernhard Sch{\"o}lkopf and Thomas Hofmann and Antonio Orvieto},
    booktitle=ICLR,
    year={2026},
}

@misc{wang2025equilibriummatchinggenerativemodeling,
      title={Equilibrium Matching: Generative Modeling with Implicit Energy-Based Models}, 
      author={Runqian Wang and Yilun Du},
      year={2025},
      eprint={2510.02300},
      archivePrefix={arXiv},
      primaryClass={cs.LG},
      url={https://arxiv.org/abs/2510.02300}, 
}

@misc{ye2025dream,
      title={Dream 7B: Diffusion Large Language Models}, 
      author={Jiacheng Ye and Zhihui Xie and Lin Zheng and Jiahui Gao and Zirui Wu and Xin Jiang and Zhenguo Li and Lingpeng Kong},
      year={2025},
      eprint={2508.15487},
      archivePrefix={arXiv},
      primaryClass={cs.CL},
      url={https://arxiv.org/abs/2508.15487}, 
}

@inproceedings{zheng2024maskeddiffusionmodelssecretly,
  title={Masked diffusion models are secretly time-agnostic masked models and exploit inaccurate categorical sampling},
  author={Zheng, Kaiwen and Chen, Yongxin and Mao, Hanzi and Liu, Ming-Yu and Zhu, Jun and Zhang, Qinsheng},
  booktitle=ICLR,
  year={2025}
}

@inproceedings{
    sahoo2026scalingmaskeddiffusionlanguage,
    title={Scaling Beyond Masked Diffusion Language Models},
    author={Subham Sekhar Sahoo and Jean-Marie Lemercier and Zhihan Yang and Justin Deschenaux and Jingyu Liu and John Thickstun and Ante Juki{\'c}},
    booktitle=ICML,
    year={2026},
}

@inproceedings{
    song2021score,
    title={Score-Based Generative Modeling through Stochastic Differential Equations},
    author={Yang Song and Jascha Sohl-Dickstein and Diederik P Kingma and Abhishek Kumar and Stefano Ermon and Ben Poole},
    booktitle=ICLR,
    year={2021},
}
